\documentclass[conference]{IEEEtran}
\usepackage{subcaption}
\usepackage[pdftex]{graphicx}
\graphicspath{{img/png/}}

\ifCLASSINFOpdf
\else
\fi

\usepackage{amssymb}
\usepackage{amsmath}
\usepackage{algorithm} 
\usepackage{algpseudocode} 
\usepackage{tabularx}
\usepackage{caption}
\usepackage{mathtools}
\usepackage{varwidth}
\usepackage[dvipsnames]{xcolor}
\usepackage{lipsum}
\usepackage{graphicx}

\usepackage{tikz}
\usetikzlibrary{positioning}
\usetikzlibrary{decorations.markings}
\usetikzlibrary{arrows.meta}

\usepackage{ctable}

\usepackage[backend=biber, style=numeric,]{biblatex}

\usepackage{caption}
\DeclareCaptionFormat{myformat}{%
  \begin{varwidth}{\linewidth}%
    #1#2#3%
  \end{varwidth}%
}
\begin{document}
\title{%Causality detection the for smart home
To do or not to do: finding causal relations in smart homes
}

%\author{\IEEEauthorblockN{Michael}
%\IEEEauthorblockA{School of Electrical and\\Computer Engineering\\
%Georgia Institute of Technology\\
%Atlanta, Georgia 30332--0250\\
%Email: http://www.michaelshell.org/contact.html}
%\and
%\IEEEauthorblockN{Homer Simpson}
%\IEEEauthorblockA{Twentieth Century Fox\\
%\and
%\IEEEauthorblockN{James Kirk\\ and Montgomery Scott}
%\IEEEauthorblockA{Starfleet Academy\\
%San Francisco, California 96678--2391\\


\author{\IEEEauthorblockN{Kanvaly Fadiga\IEEEauthorrefmark{1}\IEEEauthorrefmark{2},
Étienne Houzé\IEEEauthorrefmark{1},
Ada Diaconescu\IEEEauthorrefmark{1},
Jean-Louis Dessalles\IEEEauthorrefmark{1},
}
\IEEEauthorblockA{\IEEEauthorrefmark{1}LTCI Lab, Telécom Paris, Institut Polytechnique de Paris, Palaiseau, France \\
Email: \{first name\}.\{last name\}@telecom-paris.fr}
\IEEEauthorblockA{\IEEEauthorrefmark{2}Ecole Polytechnique, Institut Polytechnique de Paris, Palaiseau, France\\
Email: \{first name\}.\{last name\}@polytechnique.edu}}


\maketitle

\begin{abstract}
Research in Cognitive Science suggests that humans understand and represent knowledge of the world through causal relationships. In addition to observations, they can rely on experimenting and counterfactual reasoning -- i.e. referring to an alternative course of events -- to identify causal relations and explain atypical situations. Different instances of control systems, such as smart homes, would benefit from having a similar causal model, as it would help the user understand the logic of the system and better react when needed. However, while data-driven methods achieve high levels of correlation detection, they mainly fall short of finding causal relations, notably being limited to observations only. Notably, they struggle to identify the cause from the effect when detecting a correlation between two variables. This paper introduces a new way to learn causal models from a mixture of experiments on the environment and observational data. The core of our method is the use of selected interventions, especially our learning takes into account the variables where it is impossible to intervene, unlike other approaches. The causal model we obtain is then used to generate Causal Bayesian Networks, which can be later used to perform diagnostic and predictive inference. We use our method on a smart home simulation, a use case where knowing causal relations pave the way towards explainable systems. Our algorithm succeeds in generating a Causal Bayesian Network close to the simulation's ground truth causal interactions, showing encouraging prospects for application in real-life systems.
\end{abstract}

% \fi
\IEEEpeerreviewmaketitle

\section{Introduction}
\label{sec:intro}

Self-Adaptive Systems (SAS) are by nature interacting with a changing environment, be it software or physical\cite{weyns_software_2019}. In this context of interactions, the ability to model the environment is prime, as it would help to trace back failures, identifying conflicts between goals, or perform explanatory reasoning. \cite{li_explanations_2020} has shown that, in typical smart home setups, explaining decisions to the user reduces the risk of wrong interventions. However, identifying causal relations in the environment of a SAS is a hard task. % TODO ref here

Hard-coding the causal model, i.e. expressing constraints and links upon variables as rules or a static ontology is possible, but shows limited interest in the case of SAS. 
Indeed, since adaptability to a changing environment is a core feature of SAS, a static model of the environment is not suited to this configuration. Operating changes to the model could be considered, but is likely to require many human interventions, thus contradicting the principles of autonomic computing\cite{kramer_rigorous_2009}.

To illustrate this issue, consider the following situation. A user is experiencing an anomaly in the temperature control system of her smart home, as the temperature is unexpectedly cold. A hard-coded model has been implemented, which contains causal links from heater or thermometer malfunctions to the mishandling of the temperature. However, both these possibilities are discarded, as no component seems to report any problem. In this case, the cause might lie in an unexpected relation: as the user recently moved the lamp closer to the temperature sensor, and that the days, in winter, are shorter, the light is on, which produces heat, effectively skewing thermometer measures. This configuration being particular to this home, no hard-coded prior causal knowledge would be able to anticipate it without ad-hoc rules.

To overcome this common pitfall, many recent smart home systems integrate Machine Learning components to predict the environment's behavior and make optimized decisions\cite{kumar_proposal_2019}. However, spurious correlations are often found in data, especially in high dimensions\cite{fan_challenges_2014}, leading to misinterpretation and erroneous causal relations. These approaches thus mostly fall short of providing the user with a comprehensible causal model of the environment.

The theory of Causality, brought to attention among others by J. Pearl \cite{PearlMackenzie18, pearl_causality_2009}, offers tools to identify and eliminate spurious correlations in the construction of a causal model, mostly by formalizing the concept of \emph{intervention} defined by ``do-operation''. Our method consists of augmenting a standard Machine Learning approach with interventions on selected variables to infer causal relations. The result is a Causal Bayesian Network, i.e. a Bayesian Network whose structure is a causal graph of the environment.

Our approach is generic and can be applied to build causal models of various environments. But it can be computationally expensive to apply it to an environment with a very large network. We choose to apply it in the smart homes case because it offers many advantages. Firstly, we don't start from scratch as we can begin with a hard-coded causal model then incrementally improve it. Moreover, making interventions in a smart home is easier to do than in some environments (e.g. a nuclear power plant). Furthermore, the area of influence of some variables is limited to their neighborhood, which reduces the number of relationships to consider. These elements could help in scaling our approach to a home with a very large number of devices. 

Our approach's advantage is that it considers that interventions are not always possible. In ideal setups, where all variables are operable, it yields results close to ground truth causal relations. It is still able to perform well when fewer variables are available for intervention, being closer to the performance of data-driven methods.

The rest of this paper is organized as follows. In section \ref{sec:background}, we present the theoretical bases of causality and Bayesian graphs upon which our approach is built. Then, we review some existing approaches to related issues in section \ref{sec:related}. We then detail our method, and propose, in section \ref{sec:method} to illustrate it by comparing known causal graphs of a smart home with the results of our method in section \ref{sec:experiments}. Finally, we analyze the current limits of our approach and see how it can be integrated into broader systems in section \ref{sec:discussion}.

%%%%%%%%%%%%%%%%%%%%%%%%%%%%%%%%%%%%%%%%%%%%%%%%%%%%%%%%%%%%%%%%
%                           BACKGROUND
%%%%%%%%%%%%%%%%%%%%%%%%%%%%%%%%%%%%%%%%%%%%%%%%%%%%%%%%%%%%%%%%

\begin{figure}[ht!]
\centering
\small
\begin{tabular}{|m{5em}|m{3em}|m{6.5em}|m{8em}|}
    \hline
    \textbf{Level} & \textbf{Activity} & \textbf{Question} & \textbf{Example} \\
     \hline
     \scriptsize{Association $P(Y| X)$}& \scriptsize{Seeing} & \scriptsize{How would seeing $X$ change my belief in $Y$?} & \scriptsize{What is the probability of someone's presence knowing the light is on ?} \\
     \hline
     \scriptsize{Intervention $P(Y|do(X)$} & \scriptsize{Doing} & \scriptsize{What if I do $X=x$}? & \scriptsize{If I turn the heater on, would the temperature change?}\\
     \hline
     \scriptsize{Counterfactual $P(Y_x| X', Y')$} & \scriptsize{Imagining} & \scriptsize{What if things were different?} & \scriptsize{Had the light been off, would the temperature had changed?} \\
     \hline
\end{tabular}
%%%% AAAAAAAAAAHHHHHHHHHHHHHHh
%%%% OK
% je vois
% je vais voir si on peut pas la réduire en la changeant en tableau ?
% ok !
%\includegraphics[width=\linewidth]{img/png/ladder}
\caption{Pearl causality hierarchy \cite{PearlMackenzie18, bareinboim_pearls_2020}}
\label{fig:ladder}
\end{figure}

% Plan pour la section: Commencer 
\section{Theoretical background}
\label{sec:background}

Many examples from Machine Learning point out that algorithms usually lack the understanding of causal relations behind observations and predictions, causing misinterpretations of correlations\cite{obermeyer_dissecting_2019}. \cite{PearlMackenzie18} goes further by integrating this observation into a ``ladder of causation'', in which three distinct levels are identified (fig.  \ref{fig:ladder}):

\begin{itemize}
   \item \textbf{Observing} corresponds, according to Pearl \cite{PearlMackenzie18}, to the first and more reachable level of cognition: observing the world and noting correlations, dependence between some sets of variables. This stage is the ground for many modern AI approaches based on data analysis.
   \item \textbf{Acting} This advanced stage of cognition requires the agent to be able to act on some variables of the environment, observe the consequences and infer causal relations. The typical question answered at this point is ``If I do this, what will happen next ?"
  \item \textbf{Counterfactual Thinking} involves consideration of an alternate version of a past event, or what would happen under different circumstances for the same experiment. According to Pearl, this level of cognitive ability is only reached by humans. The typical question would be : "What if the apple was twice as heavy ? Would it have fallen at a different speed?"
\end{itemize}

During the twentieth century, from the causal chains of Wright \cite{wright_correlation_1921} to the integration of causal inference into Machine Learning algorithms \cite{peters_elements_2017, richens_improving_2020}, research in the field of Causality Theory aimed to formalize the intuitive concepts of cause-consequence relations.

\subsection{Structural Causal Model}

Causal models aim at representing the interactions between cause and consequence without ambiguity. From the definition of \cite{peters_elements_2017}, a Structural Causal Model (SCM) contains $C \rightarrow E$ if and only if $C \equiv N_C$ and $E \equiv f_E(C, N_E)$. That is if the cause variable $C$ can be assigned to some specific random distribution $N_C$ while $E$ can be computed from a deterministic function of $C$ and some random noise $N_E$. Note that, in this definition, both the causal relation $f_E$ and the effect noise $N_E$ are independent of the cause $C$.

\begin{figure}[ht]
\centering
\includegraphics[width=0.4\linewidth]{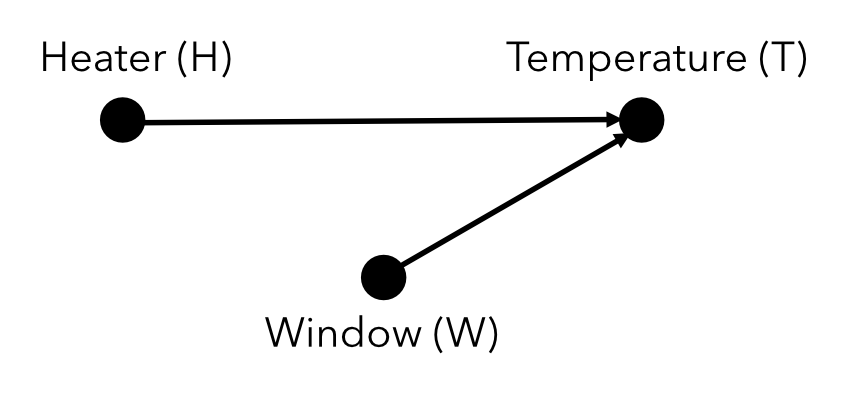}
\caption{A simple causal diagram.}
\label{fig:causaldiagram}
\end{figure}

For more complex systems, where causal dependencies between variables may be multiple, we can use a \emph{causal diagram} to represent an underlying structural causal model. A causal diagram (Fig. \ref{fig:causaldiagram}) is a directed acyclic graph \cite{PearlMackenzie18, peters_elements_2017, spirtes_causation_2000} that shows the causal relationships between variables. The nodes of the graph are the variables, and an edge $(C, E)$ belongs to the graph if and only if $C \rightarrow E$ belongs to the underlying SCM. For example, in the diagram presented in fig. \ref{fig:causaldiagram}, the arrow connecting variables \textit{Heater} and \textit{Temperature} ($H \longrightarrow T$) indicates that the temperature is causally influenced by the state of the heater. Another point of view is to consider \textit{Temperature} as listening to the \textit{Heater} variable to choose its value.

Compared to the more general definition of SCM, causal diagrams add the condition of being acyclic\cite{pearl_causality_2009}, encompassing the idea that causality flows in one direction only: if $C$ has a causal influence on $E$, then $E$ cannot influence $C$. This further prevents a variable to influence itself.

% TODO: Il faut peut être en dire un peu plus sur la considération de graphes contenant des cycles. Peut-être pointer vers quelques papiers qui le font, 

\subsection{Do-operator}

The idea of being able to intervene in the environment to test a causal relationship between variables is prime in the literature of Causality Theory \cite{pearl_causality_2009, peters_elements_2017} and can be linked to a general controlled environment experiment. 

The intervention operation has been formalized by Pearl by introducing the \emph{do-calculus}\cite{pearl_causality_2009, PearlMackenzie18}. Following his notations, $do(C = x)$ means that $C$ has been forced to take the value $x$ by an external action. It follows that, if $C \rightarrow E$ was part of the SCM, the causal relation $E= f_E(C, N_E)$ remains unchanged by this operation. This operation therefore allows to identify causal relations: if $\mathbb{P}(E) \neq \mathbb{P}(E | do(C=x))$, there is a causal connection $C\rightarrow E$. In this case, we will use the notation $do(C) \leadsto E$.

While mere observations of the variables $H, T$ and $W$ from the example of fig. \ref{fig:causaldiagram} would show correlations between $H$ and $T$, interventions would give more details on the underlying SCM. On one hand, $\mathbb{P}(Heater\mid do(Temperature=20))=\mathbb{P}(Heater)$ and on the other hand, $\mathbb{P}(Temperature\mid do(Heater=High))\neq\mathbb{P}(Temperature)$, reflecting that the heater causes the temperature change, not the other way around.

The do-operation $do(C=x)$ considers an \emph{external} intervention, meaning that it forces the variable $C$ to a given value $x$, while making it insensitive to all other variables. On a causal diagram, this is equivalent to removing all incoming edges to node $C$.  For instance, if we consider the simple causal diagram $C_0 \rightarrow C_1 \rightarrow E$, performing the intervention $do(C_1 = x)$ will remove the edge $C_0 \rightarrow C_1$, thus making both $C_1$ and $E$ independent from $C_0$, thus revealing the linear structure of the graph.

\subsection{Bayesian Network}
As Causality Theory emerged with causal diagrams, links can be made with \emph{Bayesian Networks} which are a broadly used tool for representing and modeling correlated variables \cite{koller_probabilistic_2009}. Numerous methods for training and dynamically building Bayesian Networks in many different application contexts exist in the literature\cite{bielza_discrete_2014, koller_probabilistic_2009}.

\begin{figure}[ht]
\centering
\includegraphics[width=0.6\linewidth]{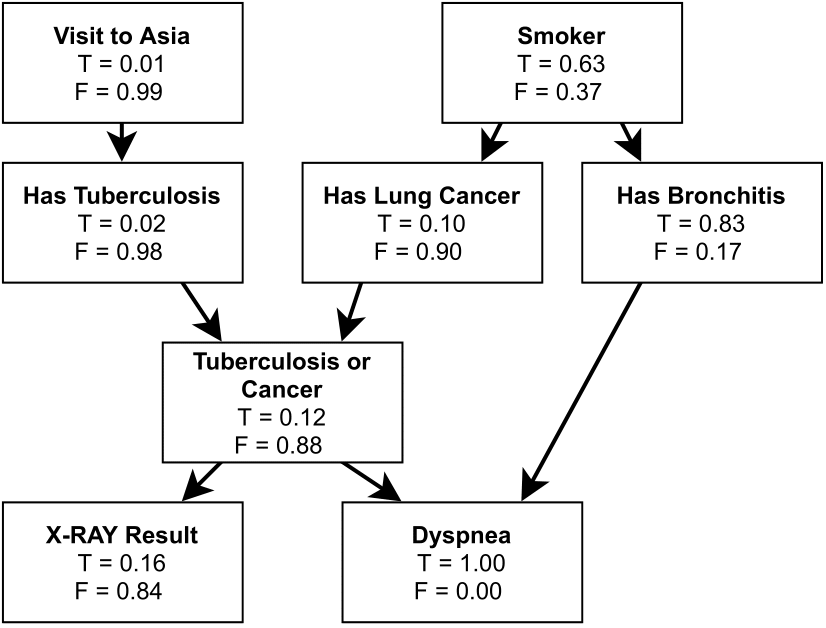}
\caption{A simple Bayesian network, known as the Asia network\protect\footnotemark}
\label{fig:bbn}
\end{figure}

Formally, a Bayesian Network (see Fig. \ref{fig:bbn}) is a directed acyclic graph (DAG) where the nodes correspond to random variables. Each node is associated with a set conditional probabilities $\mathbb{P}(X_i \mid par(X_i) )$, where $X_i$ is the variable associated with the specific node and $par(X_i)$ denotes the set of parents of node $X_i$.

To build a Bayesian network, one, therefore, needs to:
\begin{itemize}
    \item define the graph of the model, i.e. the different variables, and which ones are linked together
    \item find, for each of these variables, the table of probabilities conditioned on its parent variables
\end{itemize}

The graph is also called the "structure" of the model, and the probability tables its "parameters". Structure and parameters can be provided by experts, or calculated from data, although in general the structure is defined by experts and the parameters calculated from experimental data.

% ? Maybe add a table of difference/common points between causal diagrams and bayesian nets????

A Bayesian network carries no assumption that the arrow has any causal meaning. However, Bayesian networks hold the key that enables causal diagrams to interface with data. The probabilistic properties of Bayesian networks and the belief propagation algorithms that we will present later are indispensable for understanding causal inference.

The main differences between Bayesian networks and causal diagrams lie in how they are constructed and the uses to which they are put. A Bayesian network is literally nothing more than a compact representation of a huge probability table. The arrows mean only that the probabilities of child nodes are related to the values of parent nodes by a certain formula (the conditional probability tables) and that this relation is sufficient. That is, knowing additional ancestors of the child will not change the formula. Likewise, a missing arrow between any two nodes means that they are independent, once we know the values of their parents.

If, however, the same diagram has been constructed as a causal diagram, then both the thinking that goes into the construction and the interpretation of the final diagram change. In the construction phase, we need to examine each variable, say $C$, and ask ourselves which other variables it “listens” to before choosing its value. The chain structure $A\rightarrow B\rightarrow C$ means that $B$ listens to $A$ only, $C$ listens to $B$ only, and $A$ listens to no one; that is, it is determined by external forces that are not part of our model.

\footnotetext{From \emph{Bayes Server} (\url{bayesserver.com/examples/networks/asia})}

%%%%%%%%%%%%%%%%%%%%%%%%%%%%%%%%%%%%%%%%%%%%%%%%%%%%%%%%%%%%%%%%%
%                   RELATED WORKS
%%%%%%%%%%%%%%%%%%%%%%%%%%%%%%%%%%%%%%%%%%%%%%%%%%%%%%%%%%%%%%%%%

\section{Related Work}
\label{sec:related}
% Tian Gao et al. 2015, Local Causal Discovery of Direct Causes and Effects
% Arquimides Mendez Molina et al. 2020, Combining Reinforcement Learning and Causal Models for Robotic Applications
% Sergei Volodin et al. 2020, Resolving Spurious Correlations in Causal Models of Environments via Interventions

Over recent years, different approaches have tried to close the gap between ``classical'' observation-based Machine Learning and Causal Theory. For instance, Reinforcement Learning, as already noted by Pearl \cite{PearlMackenzie18} can be seen as a better approach than pure correlation observations, since the agent has the opportunity to act on its environment and learn from its reactions\cite{sutton_reinforcement_2018}. Thus, Reinforcement Learning has proved very powerful in tasks that were previously considered as requiring intelligent thinking, such as games\cite{schrittwieser_mastering_2020}. An agent in a reinforcement learning environment provides intervention data for learning a causal model through the exploration of the state space. However, the causal model it learns is implicit therefore it cannot be used and interpreted.

\cite{mihaljevic_learning_2018} uses a different approach to learning Bayesian Networks, trying to identify a minimal equivalence class between DAGs that fit with the observation data. The result is then presented as a Partially Directed Acyclic Graph (PDAG) which is a directed graph with undirected connections. While this method offers the advantage of keeping the graph simple and shows good predictive performance, it still only relies on mere observations, and as such lacks causal information that may impact its interpretation. \cite{causal_obs_exp} completes the previous method by orienting the undirected connections of the PDAG through interventions. He ends up orienting relations proposed by the correlations which are often non-causal relationships. Moreover, in their approach they did not take into account the fact that on some variables it is not possible to make interventions.

% cependant, le PDAG est basé que sur les corrélations, ainsi, on se retrouve avec des connections qui ne sont pas causale et aussi des relations des relations causales absentes. 

% -> Sergei Volodin et al. 2020, Resolving Spurious Correlations in Causal Models of Environments via Interventions
% --> We consider the problem of inferring a causal model of a reinforcement learning environment and we propose a method to deal with spurious correlations. Specifically, our method designs a reward function which incentivises an agent to do an intervention to find errors in the causal model. The data obtained from doing the intervention is used to improve the causal model. 
% ------> J'ai ajouté au paragraphe suivant !
% 
Some applications consider counterfactual reasoning and integrate it into the learning process of a SCM\cite{madumal_explainable_2020, volodin_resolving_2020}. In their workflow, they consider an agent that learns a causal model of its environment and then use this knowledge to perform counterfactual reasoning and improve performance. Results in providing explanations for an agent's behavior in the controlled environment of a strategy game are encouraging\cite{madumal_explainable_2020}. In a closer-to-life situation, \cite{richens_improving_2020} found that allowing do-operations in a learning framework could improve performance in a classification task and achieve better-than-humans detection of a medical condition.

Our approach differs from \cite{causal_obs_exp,mihaljevic_learning_2018}, as it accounts for some variables being unavailable to intervention. Instead, we learn causal relations using a mix of observation data and, when it is possible and useful, intervention data.

% We need something to emphasize in what ways what we do here is different from what we cite from literature: mixing bayesian and causal nets ? Advantages ???

% However many applications lack a generic approach, independent from knowledge over the application domain.

%%%%%%%%%%%%%%%%%%%%%%%%%%%%%%%%%%%%%%%%%%%%%%%%%%%%%%%%%%%%%%%%
%               LEARNING CAUSAL BAYESIAN ....
%%%%%%%%%%%%%%%%%%%%%%%%%%%%%%%%%%%%%%%%%%%%%%%%%%%%%%%%%%%%%%%%

\section{Learning Causal Bayesian Networks with Intervention}
\label{sec:method}

The base intuition for our approach is to test whether intervention on one variable $C$ has an influence over another variable $E$, observed as a change in their distribution. If so, we know from Causality Theory that there is a causal relation $C \rightarrow E$ in the SCM of the system, but an ambiguity remains whether this relation is direct or not. We therefore propose to incrementally block causal paths of nodes connected to the node on which we act, effectively narrowing down the possible relations.

We illustrate our approach in a setup consisting of Boolean variables. For illustration purposes, we consider the simple situation displayed in fig. \ref{fig:room}: a room whose temperature (hot or cold) is influenced by the state of a heater (on or off). The heater can be triggered by the user's presence in the room. Similarly, the window can be either open or close.

\subsection{Testing causal influence using interventions}

\subsubsection{Direct Influence}

Testing the direct influence between two variables $C$ and $E$ is answering the following question: \emph{``Does $C$ have an influence on $B$?''}. Our approach to this question is to incrementally remove possible causal relations following different interventions. These interventions are made by directly acting upon the environment and monitoring possibly influenced variables for changes in their probability distribution. To test possible changes, we use a chi-squared $\chi^2$ test on the distributions $\mathbb{P}(E \mid do(C=0))$ and $\mathbb{P}(E \mid do(C=1)$.

\begin{figure}[hb]
\centering
\includegraphics[width=0.35\linewidth]{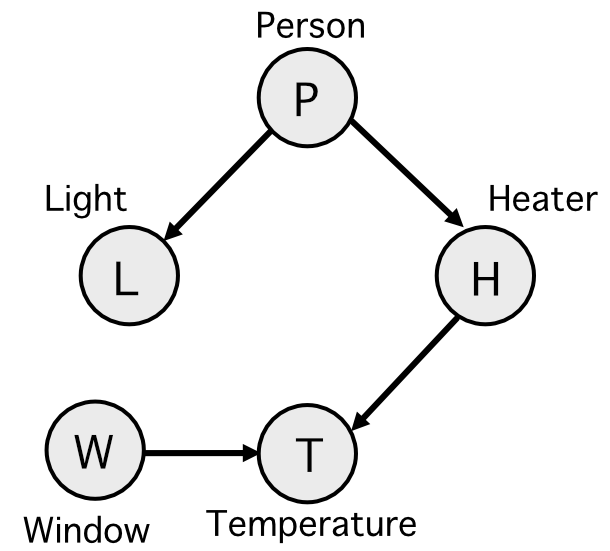}
\caption{A room causal diagram}
\label{fig:room}
\end{figure}

This test allows removing non-causal connections between pairs of variables, using both intervention operation and counterfactual reasoning. The intervention operations can be performed by directly letting our algorithm act on selected variables in the environment, thanks to the preconditions we applied to the setup. For instance, in the example of Fig. \ref{fig:room}, the distribution of L changes depending on whether the person is detected inside (P=1) the room or not (P=0). Conversely, the distribution of $P$ is not affected by the value assigned to $L$ during the intervention operation. These two operations, therefore, lead to the conclusion that $do(P) \leadsto L$ is true and $do(L) \leadsto P$ is false.

\begin{figure}[hb]
  \centering
    \begin{subfigure}{0.49\linewidth}
        \includegraphics[width=\linewidth]{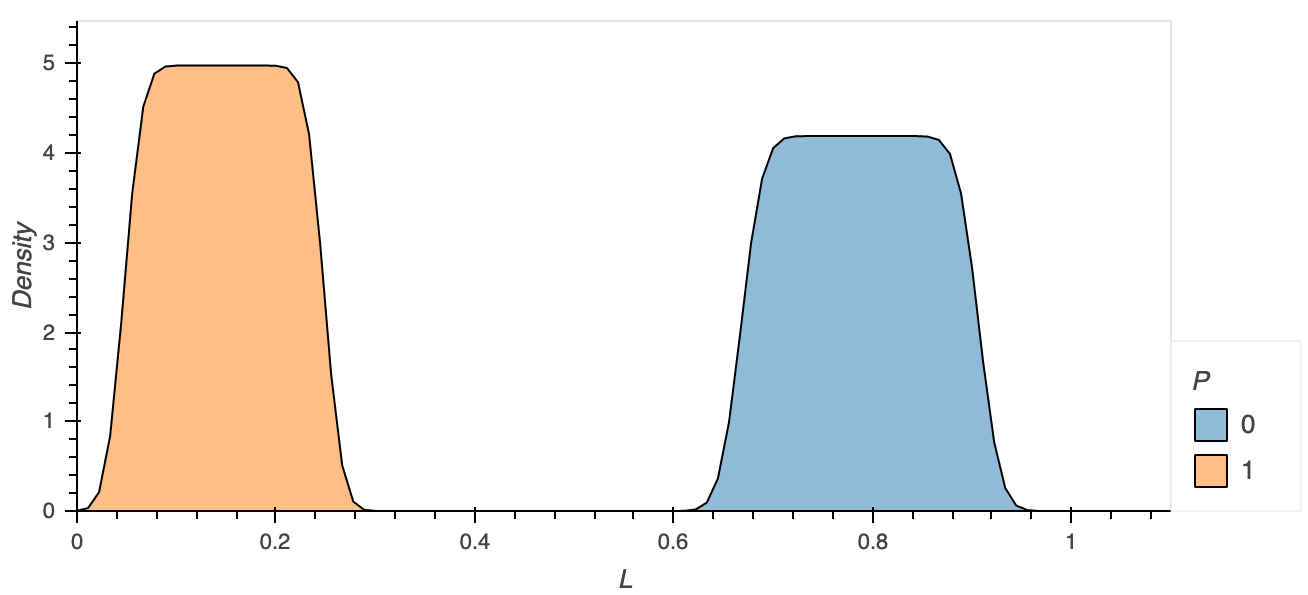}
        \caption{$do(P)\leadsto L$}
        \label{fig:PLa}
    \end{subfigure}
    \begin{subfigure}{0.49\linewidth}
        \includegraphics[width=\linewidth]{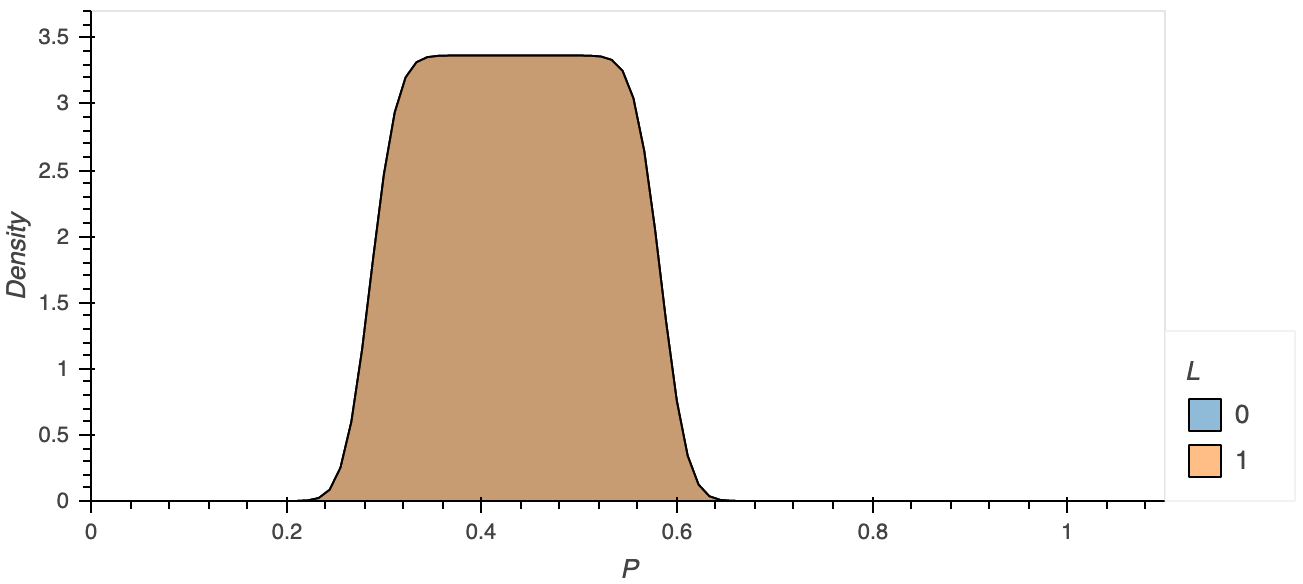}
        \caption{$do(L)\leadsto P$}
        \label{fig:PLb}
    \end{subfigure}
    \begin{subfigure}{0.49\linewidth}
        \includegraphics[width=\linewidth]{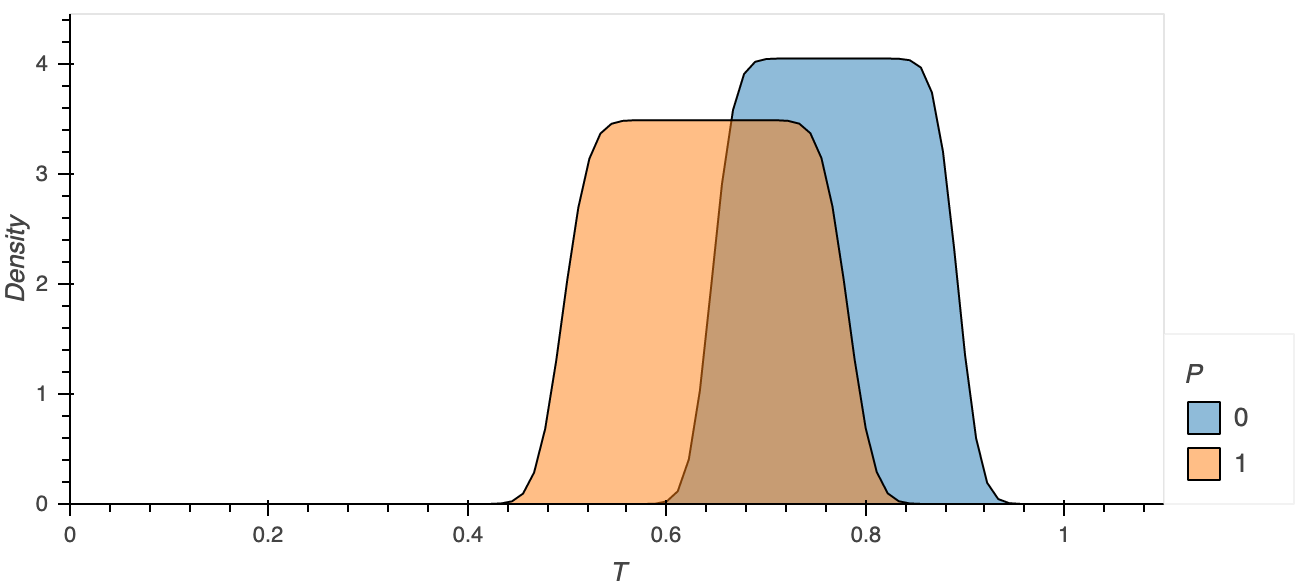}
        \caption{$do(P)\leadsto T$}
        \label{fig:PT}
    \end{subfigure}
    \begin{subfigure}{0.49\linewidth}
         \includegraphics[width=\linewidth]{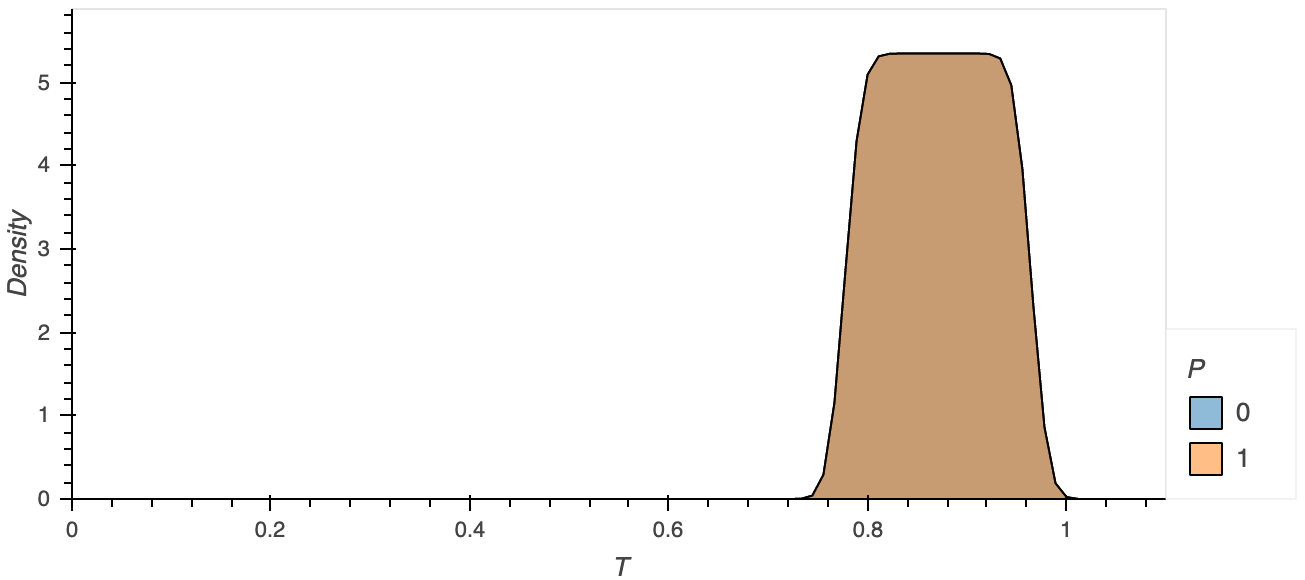}
        \caption{$do(P)\leadsto T \mid H=0$}
        \label{fig:PTcond}
    \end{subfigure}
    
  \caption{Different intervention tests. (\ref{fig:PLa}), the probability density distribution of $L$ changes depending on whether the intervention sets $P$ to 0(blue) or 1(orange). In (\ref{fig:PLb}), interventions on $L$ do not affect the probability distribution of $P$. (\ref{fig:PT}): intervening on $P$ shows a change in the distribution of $T$. However, conditioning this relation with $H=0$ removes the relation(\ref{fig:PTcond})}
  \label{fig:PL}
\end{figure}

\hspace{1cm}
\subsubsection{Conditional Influence}

\begin{figure*}[ht]
    \centering
    \begin{subfigure}{0.18\linewidth}
        \centering
        \includegraphics[width=\linewidth]{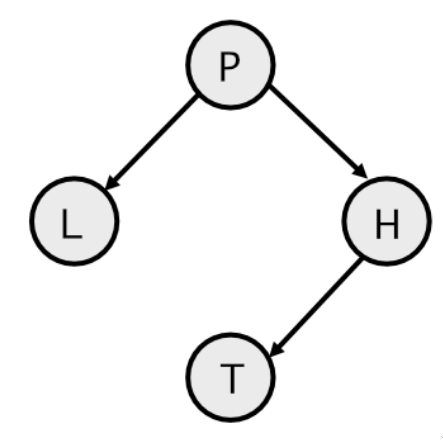}
        \caption{}
        \label{fig:algo_gt}
    \end{subfigure}
    \begin{subfigure}{0.18\linewidth}
        \centering
        \includegraphics[width=\linewidth]{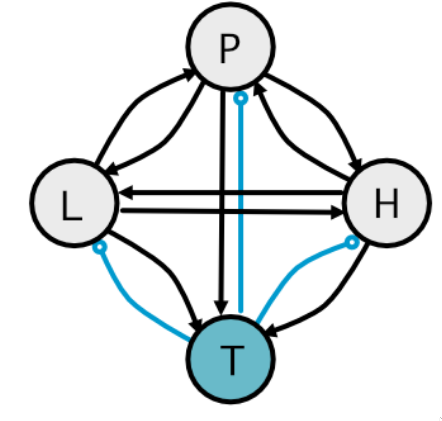}
        \caption{}
        \label{fig:algo_1}
    \end{subfigure}
    \begin{subfigure}{0.18\linewidth}
        \centering
        \includegraphics[width=\linewidth]{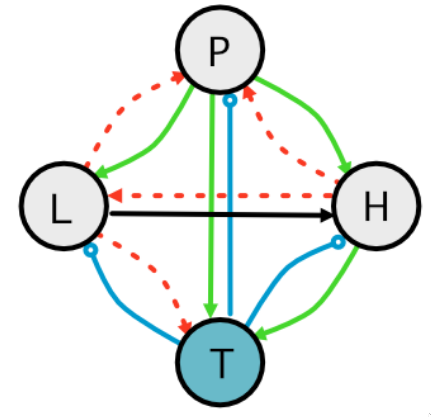}
        \caption{}
        \label{fig:algo_2}
    \end{subfigure}
    \begin{subfigure}{0.18\linewidth}
        \centering
        \includegraphics[width=\linewidth]{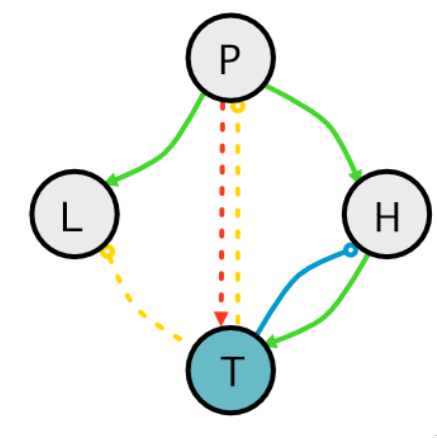}
        \caption{}
        \label{fig:algo_3}
    \end{subfigure}
    \begin{subfigure}{0.18\linewidth}
        \centering
        \includegraphics[width=\linewidth]{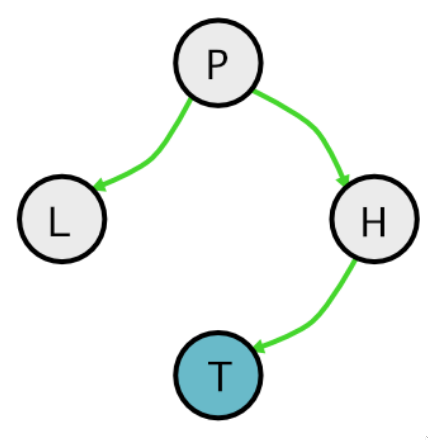}
        \caption{}
        \label{fig:algo_4}
    \end{subfigure}
    \caption{Principle of our algorithm. (\ref{fig:algo_gt}) ground truth causal model of the environment. (\ref{fig:algo_1}) Initialization to a fully-connected graph over the variables. Non-doable arrows and nodes are shown in blue. (\ref{fig:algo_2}) Causality tests with interventions remove the red arrows. (\ref{fig:algo_3}) Arrows are removed, either by independence test (in yellow) or causality test (in red). (\ref{fig:algo_4}) The final graph is obtained by removing cycles, prioritizing non-doable arrows.}
    \label{fig:algo_exe}
\end{figure*}

The case of evaluating a conditional causal influence can be summarized with the question: \emph{``Did $C$ have an influence on $E$ given $B$?''}($do(A)\leadsto C \mid do(B)$). As opposed to the previous case, the causal path is indirect and thus requires additional processing. Here, we process by testing if the causal influence between $C$ and $E$ still holds conditioned on the value of the third variable $B$. That is, checking if, for some value $u$ taken by $B$, we would observe, via a chi-squared test, a difference between $\mathbb{P}(E \mid do(C=0), do(B=u))$ and $\mathbb{P}(E \mid do(C=1), do(B=u))$.

This operation can be viewed as ``locking'' the value of $B$ to a given value $u$, and observe if the causal relation holds. In the examples of  fig. \ref{fig:PT} and \ref{fig:PTcond}, we infer the causal relations:  
\begin{itemize}
    \item $do(P)\leadsto T \mid do(L=0)$ : True
    \item $do(P)\leadsto T \mid do(H=0)$ : False
\end{itemize}

\subsection{Causal Learner Algorithm}

Our algorithm, presented in alg.\ref{alg:algo}, iterates over the previously described elementary steps to remove non-causal pairs of variables. To this end, we start by considering a fully connected graph over all the variables in the system (see fig. \ref{fig:algo_exe}). Then, selected causal influence tests are used to remove arrows for unrelated variables. These tests are performed by increasing order of conditioning: this allows to test the costlier high-order conditioned influences on few arrows, as many have already been discarded by the first series of tests.

As shown in fig. \ref{fig:algo_exe}, a major limitation of this approach is that some do-operations are not feasible in realistic setups: in our example, this is the case for the temperature variable $T$, as one does not arbitrarily set the temperature of a room to some fixed value without modifying other variables (e.g. the heater state $H$). We therefore call the corresponding temperature node a \emph{non-doable node}, and consider all of the outgoing relations as \emph{``non-doable arrows''}, or ND-arrows, in the graph. These arrows are not directly removable since the corresponding do-operations cannot be performed.

\begin{algorithm}[ht]
    \begin{algorithmic}[1]
    \small
        \State \textbf{Initialization}: \\$\quad\mathcal{G}$ is the fully connected graph over nodes of $\mathcal{X}$
            \\ $\quad k \leftarrow 0$
        \While{There are nodes with more than $k$ neighbors in $\mathcal{G}$}
            \For{each such node $X_A$, each of its neighbors $X_B$}
                \For{each subset $\mathcal{S}$ of $k$ neighbors of $X_A$}
                    \State{\emph{\# Influence test for doable node}}
                    \If{$X_A$ is doable}
                        \State Compute $do(X_A) \leadsto X_B | do(\mathcal{S})$ 
                        \State Remove $A \rightarrow B$ from $\mathcal{G}$ if need to be
                    \State \hspace{-\algorithmicindent} \emph{\# Independence test for non-doable node}
                    \Else
                        \State Compute $Corr(X_A, X_B | \mathcal{S})$
                        \State Remove $X_A \rightarrow X_B$ if variables are independent
                    \EndIf
                \EndFor
            \EndFor
            \State $k \leftarrow k+1$
        \EndWhile
        \State Postprocess $\mathcal{G}$ to turn it into a DAG by removing least significant arrows.
    \end{algorithmic}
    \caption{Extended \emph{do} Causal Learning Algorithm}
    \label{alg:algo}
\end{algorithm}

% One limitation of the previous steps is that it assumes that one can intervene (do-operation) on all the variables. In a simulation it may be possible for all variables but in real life that might be not always possible. We won't be able to make a do-operation on whether it's the day or the night even if we wanted to. So we introduce the notion of \textbf{non-doable node (ND-node)} which are the variables we cannot intervene. For non-doable node it will no longer be possible to remove the outgoing arrows using the previous methods. We call these arrows \textbf{non do-decidable arrow (ND-arrow)}.
\begin{figure}[ht]
\begin{center}
\includegraphics[width=\linewidth]{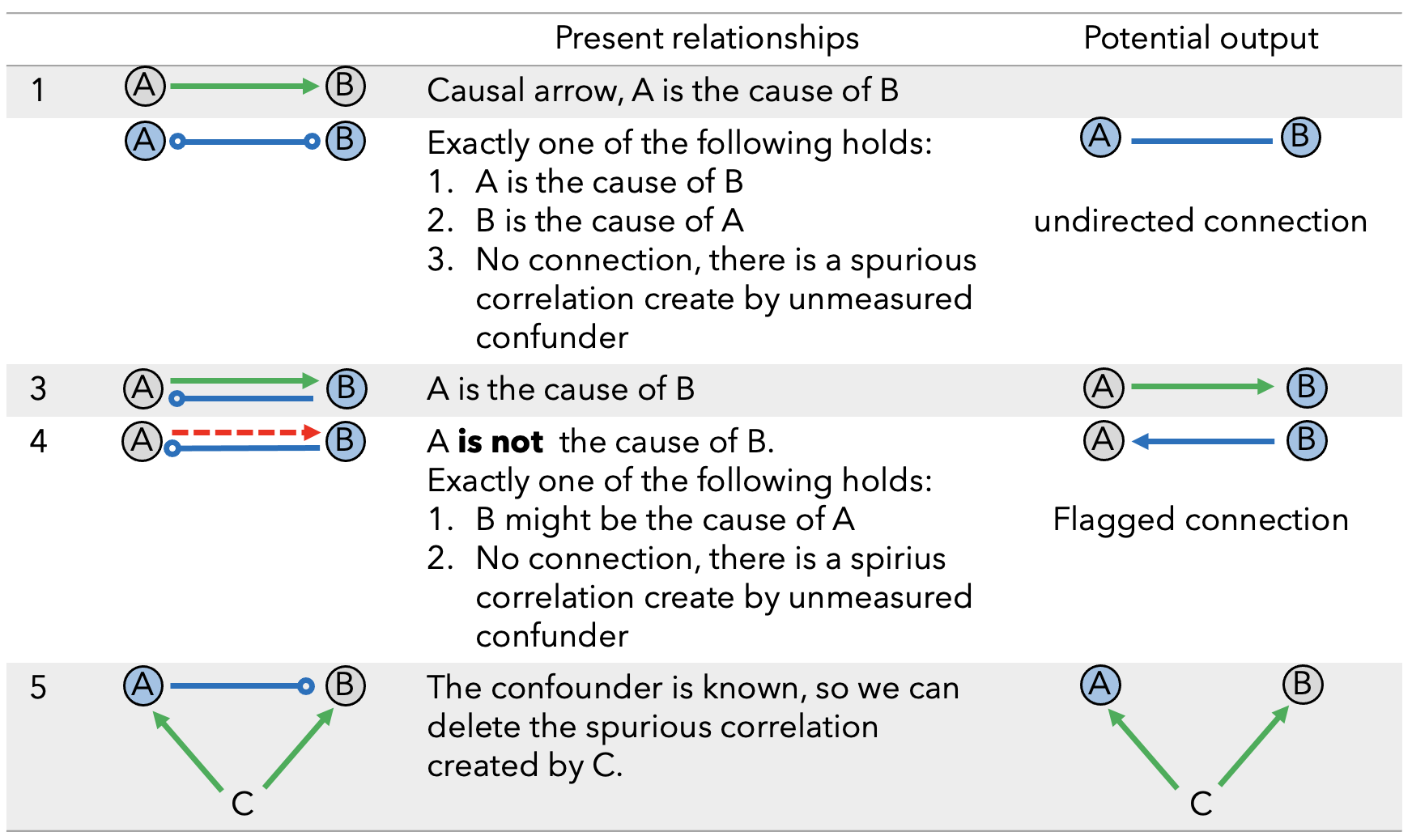}
\caption{The different possible configurations for processing the remaining ND-arrows. Causal arrow are shown in green, ND-arrow are in blue. Red arrow represent arrow remove by causality test}
\label{fig:situation}
\end{center}
\end{figure}
Processing ND-arrows, therefore, requires another approach. First, similarly to the PC-algorithm from \cite{spirtes_causation_2000}, we use a simple chi-squared test to identify whether the two variables are correlated since a causal relationship implies a correlation between variables. This first step allows to remove some connections, but, for the remaining connections, it does not provide any direction for the relation. Furthermore, one needs to be cautious about the potential risk of mislabeling correlations as causal relations. As such, remaining ND-arrows should be considered only as candidate causal relations.

% For ND-arrow we will use observational data to decide of we keep them. We use chi-square conditional independence test used in data-driven methods. If the two variables are independent the test will remove the connection. This test is used in constraint methods for learning Bayesian Network structure (PC algorithm \cite{PC}). In PC algorithm, this test is used to build an undirected graph that will be oriented later. Thus, it have no sense of direction. An important thing to remember is that this test can be misleading by correlations and therefore its results should be taken with caution.

To handle the rest of the process, we rely on the fact that the causal diagram is, by definition, a DAG. This condition leads to the removal of some arrows among the remaining candidates. Depending on the configuration, different possibilities are considered, as fig. \ref{fig:situation} shows:

% The last thing to do now is to make our algorithm return a DAG. To do this, we will analyze the different cases that we might have in the output graph of our algorithm. 

\begin{itemize}
    \item \textbf{case 1}: As no ambiguity exists, the arrow is kept in the graph.
    \item \textbf{case 2}: No information can be gathered through correlation study alone. If no direction creates a cycle in the graph, the algorithm will keep the undirected relation, and tag it as potentially spurious. Further data may lead to eliminating both of the arrows.
    \item \textbf{case 3}: One direction of the relation has been tested through a do-operation, while the other has not. The algorithm will therefore keep the direction that has been tested with an intervention. % due to the ND-nature of the left node. As a result, the algorithm will automatically discard the ND-arrow, as keeping it would create a cycle with a decided arrow. %% kanvaly : we discard also when the cond. intependence 
    \item \textbf{case 4}: While this ND-arrow can be a spurious correlation, the algorithm will keep it if it does not create a cycle in the resulting graph. It will however be flagged as such. Otherwise, the arrow is removed from the graph. More generally, if keeping several ND-arrows would lead to a cycle, the algorithm will remove the least significant one with respect to Chi-square score. % ! TODO: what measure is used for this ? Chi-squared ? Correlation value ? -> Chi-square

    %% ->kanvaly : avant on avait (elimnation of one or both) mais c'est plutot (both directly) parce que c'est pareil dans les deux sens
    %% -> Etienne : OK!
    
    \item \textbf{case 5}: In this case we see an ND-arrow that can be preserved if there is a confounder that creates a correlation between A and B. Here we see that C is a confounder. So we drop the ND-arrow.
    %In this case, the algorithm will drop the least significant arrow to remove the cycle. Since ND-arrows could not be tested with interventions, the algorithm prioritizes their deletion.
    % ! Is this really possible ? If both arrows are green, I think it means both passed the do-operation tests, so both should be true causal relations? SO this could not happen? Or do you consider the case, to make it more robust to potential errors by the algorithm??
    %--> in our case it is not possible. but if there is retroaction it can happen -> OK !
\end{itemize}

% With this analysis we have identified the following rules:

% \begin{itemize}
%    \item If there is two non decidable arrow between two nodes we replace them by a undirected connection
%    \item remove all non decidable arrow that might create loop with the causal arrow network (green arrows)
%    \item remove the least significant causal arrow in loops.
%\end{itemize}

% Because the algorithm does not necessarily return an acyclic graph, we will do cycle detection and drop some connections by removing arrows with the least significant causal influence in each cycle.

After processing all ambiguous cases, the algorithm outputs a DAG representing a causal model compatible with observations from the system. This causal diagram can then be used as a basis for further analysis. A first possibility is to use it to infer potential causes to unusual situations, and as such be included in a broader-scoped explanation process\cite{houze_decentralized_2020}. A second prospect, detailed here, is to use this diagram as the structural basis of a Bayesian Network for finer causal inference.

Our algorithm requires an exponential number of tests with respect to the number of variables (node). But, as mentioned in \ref{sec:intro}, in the case of the smart home, we can consider an incremental approach. A base, generic, causal model is implemented, which is later refined testing its connections with our method. Then, as new devices are added to the system, we consider only the newly created interactions, thus incrementally augmenting the causal graph.

\subsection{Causal Model to Bayesian Network}

In the literature, training a Bayesian is usually divided into two main parts\cite{koller_probabilistic_2009}: learning the structure of the graph and estimating its parameters. Since we use the previously learned causal diagram as a base structure, we will only focus in this part on learning the different parameters of the network, i.e. the probability tables for each node conditioned on its parents. We will call the resulting graph a \emph{Bayesian Causal Network} to emphasize its particular structure: while usual Bayesian Networks do not entail causality between their nodes, our approach leads to a graph whose connections entail a cause-effect relation.

% The training of a Bayesian network consists of two main parts: estimating the parameters of the model and learning the structure of the network
%. Here we will use the causal diagram that we learn previously. That why we can call our Bayesian Network : Causal Bayesian Network. So what's remain now is to learn the model parameters. Our parameter correspond to conditional probability table entry.

\begin{figure}[h]
\begin{center}
\includegraphics[width=0.7\linewidth]{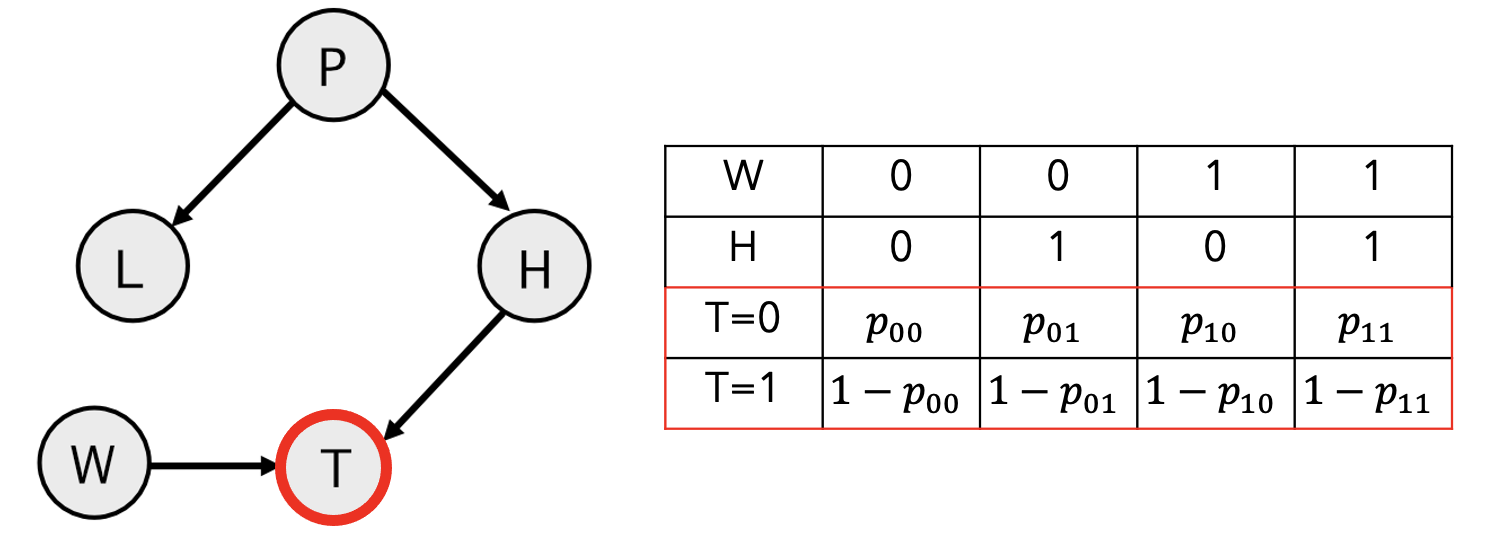}
\caption{Example of a small Bayesian network. The probability table for node T is displayed.}
\label{fig:param}
\end{center}
\end{figure}

To estimate these parameters, a conventional approach is to use a maximum likelihood estimator\cite{BNeng}, which can be resumed as estimating variables values given their parents' values only from past observational data. For example, if we consider the graph from fig. \ref{fig:param}, we would compute $p_{00}$ with:

\begin{equation}
    p_{00} = \frac{N_{T=0, (W,H)=(0,0)}}{N_{T=0, (W,H)=(0,0)} + N_{T=1, (W,H) = (0,0)}}
\end{equation}
where $N_{T=i, (W,H)=(j,k)}$ is the number of past occurrences of $(T,W,H) = (i,j,k)$.

%Maximum likelihood estimator shows that parameter learning can be resume to counting variable value given their parent value and compute a probability from it. We improve this counting process with our do-operator. Let's take a simple example to illustrate how we improve this learning through our data augmentation.

% Classical methods use only historical data. if we want to compute $p_{00}$ we will use:

% \[p_{00}=N_{T=0,WH=00} / (N_{T=0,WH=00}+N_{T=1,WH=00})\]

However, this conventional approach might be facing some issues for some estimations, notably if the number of occurrences is small. For instance, in our small example of fig. \ref{fig:param}, it might not be possible to estimate the parameters $p_{01}$. The introduction of the do-operation removes this limitation, as it becomes possible, for doable nodes, to generate all kinds of situations required to observe the outcome and estimate missing parameters of the Bayesian Network.

% with $N_{T=0,WH=00}$ the number of time $T=0$ given that $WH=00$. One issue with using historical data is that there may not be certain cases in the data or only flew. For example, if W never takes the value 1 in our data, we will not be able to estimate the value of $p_{10}$ and $p_{11}$ [\ref{fig:param}]. To solve this issue we generate suffisant data using our do-operate to make these estimation possible. For each case, let's say $W=1, H=0$ we will use N time our do-operator on the simulation and count the outcame of T. Then we will use it to compute $p_{10}$. We only use this method when there is not many observational data.

% A Bayesian network learned from the data is not necessary causal. Here our structure comes from a causal graph so what we will be manipulating is a \textbf{Causal Bayesian network (CBN)}.

\subsection{Causal Inference on Causal Bayesian Network}

% ! Pour cette partie, je pense qu'il serait bien de voir, dans les related works, des applications aux réseaux bayésiens (prédiction, diagnostic). Comme ça, on peut reprendre ici les exemples déjà abordés, et dire en quoi ce que l'on présente est complémentaire/mieux/différent. Cela permet de mieux ancrer ce que l'on dit dans les travaux existants, et fournit du contexte et des points de comparaison.

Upon completion of the training, the resulting Causal Bayesian Network can be seen as a ``conditional probability machine''\cite{koller_probabilistic_2009}. It can be used for different tasks requiring inferring new knowledge on the system. For instance, \cite{awarebn} shows how a Bayesian Network can be used to compute the probabilities of different activities compatible with the observed interaction of a user with IoT device in a smart home. This example shows the different possibilities offered by a Bayesian Network: diagnostic and predictive inference.

\begin{figure}[ht]
  \centering
    \begin{subfigure}{0.32\linewidth}
        \centering
         \includegraphics[width=\linewidth]{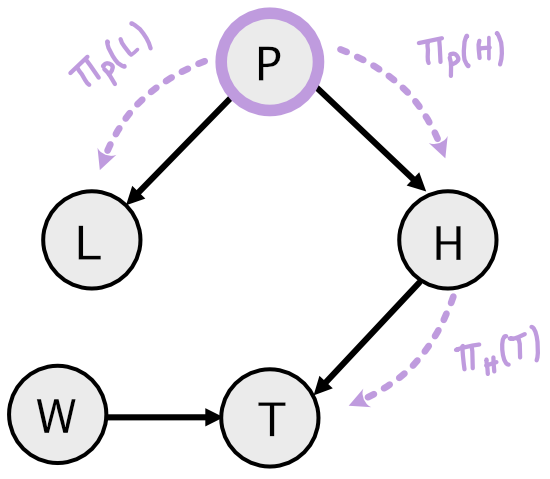}
        \caption{Predictive}
        \label{fig:pred}
    \end{subfigure}
    \begin{subfigure}{0.32\linewidth}
        \centering
         \includegraphics[width=\linewidth]{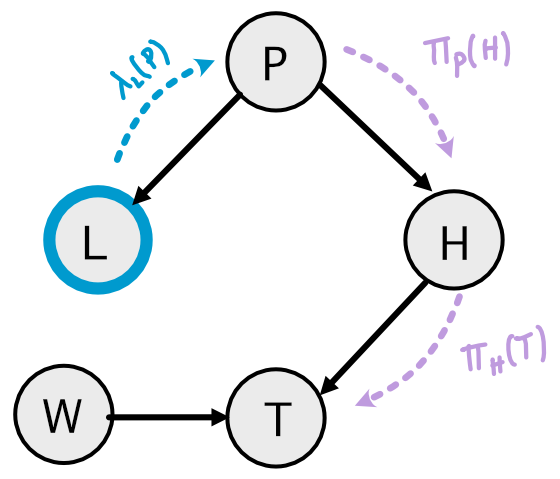}
        \caption{Diagnostic}
        \label{fig:diagnostic}
    \end{subfigure}
    \begin{subfigure}{0.32\linewidth}
        \centering
        \includegraphics[width=\linewidth]{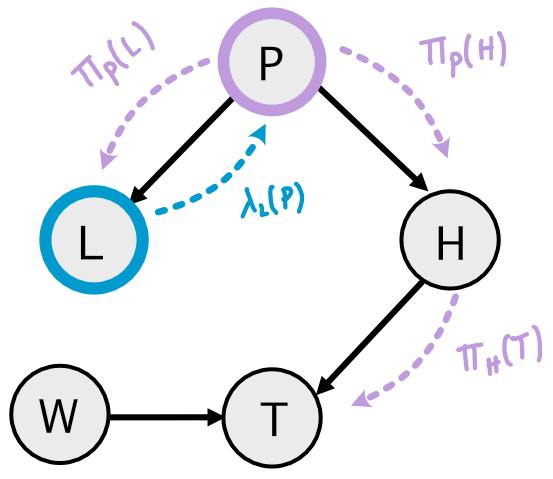}
        \caption{Both Directions}
        \label{fig:both}
    \end{subfigure}
    
  \caption{Belief propagation in a Bayesian Network can be either forwards (\ref{fig:pred}) for predictive applications, backwards (\ref{fig:diagnostic}) for diagnostic purpose, or both-oriented (\ref{fig:both}).}
  \label{fig:belief}
\end{figure}

\begin{itemize}
    \item \textbf{Predictive}: This kind of inference is interested in ``guessing'' the most probable future state of the system, given a configuration, i.e. answering the question: \emph{``What happens if $X$ is equal to $x$?''} As fig. \ref{fig:pred} shows, if evidence is put on node $P$, the inference will propagate following the direction of causal arrows, to the children of the affected node $P$.
    \item \textbf{Diagnostic}: On the other hand, diagnostic inference is interested in looking into the probable causes of observed consequences: \emph{``what would be the cause of $X=x$?''} The inference therefore goes backwards, as displayed in fig. \ref{fig:diagnostic}: from the observation on $L$, we infer the probable state of $P$, which will entails consequences over $H$ and $T$.
\end{itemize}

% Pour commencer il faut la probabilité à priori pour chaque noeud du graph. Nous allons ensuite l'utiliser pour calculer la probabilité à posteriori. à chaque n
% Essai de réécriture de la partie sur l'inférence
In either case, inference works as follows: we denote by $Bel(X=x)$ the belief that a node takes a given value (see fig. \ref{fig:bbn}, where beliefs are displayed for each node).  Following observation of the system, the beliefs of one or several nodes are set to a set value. For instance, in fig. \ref{fig:param}, knowing that the person is present will set the value of $P$ to $1$ with a probability $1$. This change of beliefs will then be propagated through the graph, following Bayes' rule.

% Ainsi à tout moment, chaque état de chaque variable sera décrite par $Bel(X=x)$ qui est la probabilité que X soit dans l'état x. Sur la figure 3 on peut voir un à quoi cela ressemble pour un reseau complet. A chaque fois qu'on a une nouvelle information sur un noeud, l'information peut être propagé localement entre les noeuds du graph. Cette propagation agit par transmission de messages entre noeuds voisins, transitant par les arcs entre ces noeuds. Le but étant que chaque noeud apprenne toute l’information et fasse connaître à l’ensemble du graphe l’information élémentaire qui le concerne. $\lambda(X=x)$ ou $\lambda(x)$ est le cumul de toutes l’information qu'un noeud à reçu de ses enfants pour la croyance au fait we $X=x$. On appellera $\lambda_U(x)$ la contribution de l’enfant U dans $\lambda$ . De même pour $\pi(x)$ qui est l'information des noeuds parents avec $\lambda_V(x)$ la contribution du parent V dans $\pi$.  

While we let the details of the propagation algorithm out of the scope of this paper (readers interested in a complete description of the process may refer to \cite{pearl_fusion_1986, koller_probabilistic_2009}), we could visualize the propagation mechanism as follows. The propagation algorithm is iterative. At every step, each node $X$ passes the following messages: to its children $Y$, $\pi_X(Y)$ containing transition probabilities; to its parents $Z$, $\lambda_X(Z)$ containing likelihood information. Conversely, it receives messages $\pi_Z(X)$ from its parents, and $\lambda_Y(X)$ from its children (fig.\ref{fig:belief}). Each node then updates its beliefs according to the messages it receives: 
\begin{equation}
    Bel(X) = \alpha \lambda(X) \pi(X)
\end{equation}
where $\alpha$ is a normalizing factor, $\lambda(X) = \prod_{Y} \lambda_Y(X)$ and $\pi(X) = \prod_{Y}\pi_Y(X)$ are the products of all messages received from children and parents, respectively. As shown by \cite{pearl_fusion_1986}, for DAGs, this propagation method converges to the belief values satisfying the observations and the network's parameters in a finite number of steps.

The predictive and diagnostic inference then allows answering various queries about the environment without having to further intervene on the system. Applications of such knowledge are further discussed in sec. \ref{sec:discussion}. One might however note that, as opposed to a traditional Bayesian Network, our proposed CBN uses only causal relations. As such, one might argue that the entailed reasoning appears more ``natural'', a case confirmed by the observation that causal relations are algorithmically simpler \cite{peters_elements_2017, janzing_causal_2010}.

% TODO: revoir cette partie, ce n'est pas clair du tout !

% The prior probability for each node needs to be calculated before the reasoning process is performed. Posterior probability or belief of variable $X=x$ is denoted as $Bel(x)$.The types of propagation of evidences that influence variable X are split into two $\pi(x)$ and $\lambda(x)$. $\pi(x)$ indicates that the evidence passes through the arrow between X and its children, and $\lambda(x)$ indicates that the evidence propagates from X to its parents. The belief, $Bel(x)$, can be written as in Eq.(\ref{eq:belief})below.

% 

% Once we have learned the Bayesian causal network, we will be able to answer many questions about the environment without even intervening. This is because we have a causal diagram packaged in a Bayesian network. The causal model gives the causal relationship and the bayesian network property can do propagation of evidence of a variable which allows an update of the probability distribution of the other variables in the network in the light of the newly found evidence. With this model we will perform two kind of operations :

% ! Table 1 and 2 have the same values!!!!!
\begin{table}[hb]
\parbox{.45\linewidth}{
\centering
\begin{tabular}{ | m{1cm}  m{2cm}| } 
    \hline
    node & ( x=0, x=1 )  \\ 
    \hline
    P & (0.5, 0.5) \\ 
    L & (0.45, 0.55) \\ 
    H & (0.7, 0.3) \\ 
    T & (0.67, 0.33) \\ 
    W & (0.7, 0.3) \\ 
    \hline
    \end{tabular}
\caption{Prior probability of each node of \ref{fig:diagnostic}}
}
\hfill
\parbox{.45\linewidth}{
\centering
\begin{tabular}{ | m{1cm}  m{2cm}| } 
    \hline
    node & ( x=0, x=1 )  \\ 
    \hline
    P & (0.89, 0.11) \\ 
    L & (1, 0) \\ 
    H & (0.86, 0.14) \\ 
    T & (0.73, 0.27) \\ 
    W & (0.7, 0.3) \\ 
    \hline
\end{tabular}
\caption{Posterior probability after L=0.}
}
\end{table}

\section{Experiments and results}
\label{sec:experiments}

\subsection{General Workflow}

As previously stated in sec. \ref{sec:intro}, we apply our methods to a smart home environment. This choice is motivated by various reasons. First, smart homes provide good examples of closed environments managed by SAS. As such, they also provide simulators, which can be used to implement an intervention operator without being limited by common physical constraints (time, safety issues, incompatibilities). In addition, they can present unusual or surprising situations where the use of a causal diagnostic can help intervene on the system to improve performance\cite{li_explanations_2020, houze_decentralized_2020}. The following section describes in detail our workflow, from data generation to training the CBN and using it for inference tasks. 

% As explained in the introduction, the smart home framework is relevant for evaluating our model.  We will present the environment in which we will work and then present different scenarios that show the usefulness of having a method like ours.

%We have evaluate our approach on two environments. The first one is electronic logic gate circuit and the second one is a simulation of smart home. Electronic gate circuit was a minimalistic scenario use to valid our method. The smart home simulation was a more complex environments  where we explore many scenarios to see our this causal learning methods can help.

%\subsection{Logical gate circuit}

%To evaluate our algorithm, we tested it on the learning of causal relations on a logic gate circuit. Here were are dealing with an evironment where all nodes are doable. The first environment is a logic gate circuit [Fig. \ref{circuit}] where sensors are placed at the input and output of each logic gate. Then to add a layer of difficulty we tested our algorithm on a second environment where the sensors are only placed at few location to create hidden variable. 

%\begin{figure}[H]
%  \centering
%  \label{P_T_cond}\caption{logical gate circuit}
%    \subfloat[circuit]{\label{circuit}\includegraphics[width=45mm]{img/png/circuit.png}}
%    \subfloat[circuit causal diagram]{\label{circuit-causal}\includegraphics[width=45mm]{img/png/complete.png}}
%\end{figure}

%Our method was able to recover all the connection of the circuit. It also recover the correct causal diagram with hidden nodes.

% TODO : resultat des autres méthodes d’apprentissage 

\subsubsection{Smart Home simulator}

\begin{figure}[ht]
\begin{center}
\includegraphics[width=0.6\linewidth]{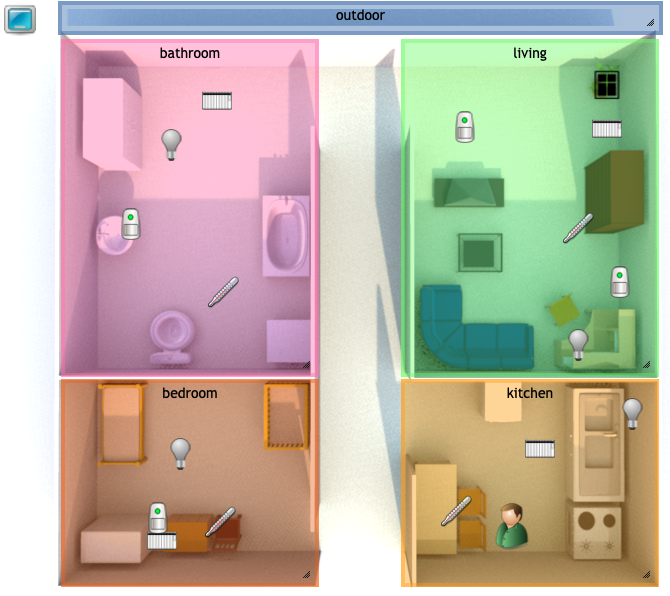}
\caption{The iCasa GUI showing the basic setup for our experiments: four rooms equipped with a presence sensor triggering heating and lighting, and a thermometer.}
\label{fig:icasa}
\end{center}
\end{figure}

Our experiments are built upon the iCasa platform \cite{lalanda_service-oriented_2020}. Based on the OSGi framework, it offers a service-oriented platform for simulations of smart home physical systems. Its autonomic manager keeps track of currently used devices, which allows for runtime deployment and modification of configuration. This enables the simulation of scenarios where variable interactions are more intricate. In our example, we simulated the behavior of different rooms, each one characterized by physical variables such as temperature and illumination. Each of these rooms is equipped with different devices able to monitor or modify the room's physical variables: heater, thermometer, presence sensor, light, etc. Table \ref{tab:variables} shows the different monitored variables of the example. The entire configuration is shown in fig. \ref{fig:icasa} using the iCasa Web UI.

Using a simulation, as opposed to using a real setup, brings two main advantages for our experiments. First, it allows having a perfect knowledge of the ground-truth causal interactions, as they are directly encoded into the simulator. Secondly, it provides easy control over different parameters and thus allows to perform, if desired, some interventions that would not be feasible in real life. This will allows us to test the effect of having access to more or less possible interventions for our algorithm.
\begin{table}[]
    \centering
    \begin{tabular}{|c|c|}
        \hline
        \small
        Boolean variable & Simulation measure \\
        \hline
        person $P$ & (User.x, User.y) $\in$ room \\
        outdoor $O$& outdoor.temperature $\ge$ threshold \\
        light $L$& light.powerStatus $= 1$ \\
        presence $Pr$& sensor.presenceSensed $= 1$ \\
        power $Pow$& house.powerConsumption $\ge $ threshold \\
        thermometer $T$ & thermometer.temperature $\ge$ threshold \\
        window $W$& window.open $= 1$\\ \hline
    \end{tabular}
    \caption{Correspondence between simulation measures and the Boolean variables.}
    \label{tab:variables}
\end{table}

\subsubsection{Observation Data Generation}

Once the initial setup is complete, we let the simulation run while the various devices are left in ``autonomous mode'', i.e. they are able to adapt to changes to maintain some key environment variables within a target range, for instance, the temperature and CO2 concentration of each room. At runtime, we randomly act on some of these variables or components to observe how the system reacts to change. In total, our continuous observation generated around 500 data points. Since values from variables are originally numerical, we convert them to Boolean values by using simple threshold comparisons. Thus, we obtain a set of Boolean observations which are used to observe correlations between variables.

% We start by generating the observational data. For that we let the house in autonomous mode, which mean that the automatic controllers are activated. These controllers allow the house to maintain the temperature, ventilation, lighting and others variables at a certain level. We activate them while collecting the data because in reality it is in this way that we will collect the observational data. Thus we have randomly generated 5,000 data points.

\subsubsection{Intervention Data Generation}

To perform interventions, we use the possibility offered by the simulator setup to disable some devices. Disabled devices will no longer react to their input sensors, thus achieving the Markov blanket independence implied by intervention\cite{pearl_causality_2009}. Then the value of the device is set to a fixed value. For instance, the intervention $do(Heater=1)$ will cause the heater to turn on while being insensitive to any environmental factor such as the detection of the user's presence.

% ! question: l'état dans lequel on prend la maison est-il l'un des états générés par la méthode plus haut ? -> non non
To generate intervention data, we then proceed as follows: we sample the house in a state $s$ where each variable is assigned a value, and from this state, make an intervention $do(X=x)$ on a selected variable. We did 20 interventions per node at each stage. After a set time period $\Delta_t$, we measure the resulting state of the house $s'$, eventually considering only variables of interest (variables correlated to $X$). The period $\Delta_t$ is set to a given value manually chosen from prior experiments with the simulator, to allow the system to reach an equilibrium state after the intervention. We will discuss further time considerations in sec. \ref{sec:discussion}. 

\subsection{Results}

% ! TODO compléter ici avec les réusltats d'apprentissage et les réseaux obtenus, les comparer avec le réseau connu de la maison.

The causal model of the simulation we used for our experiments is displayed in fig. \ref{fig:gt}. We first consider three variables, namely the presence sensor, the house's power consumption, and the room's temperature as ND-nodes. While the simulation setup would allow us to intervene on them, we introduced this limitation to observe the impact of ND-nodes on causal discovery.

% Our smart house has as causal model the structure of figure 11. In this example we have considered that three variables are ND-node: the presence detector (presence), the power consumed (power) and the temperature sensor (thermometer). We will not be able to make any interventions on them, therefore, we will use the observational data to find their influence. For the rest of the variables we will be able to make interventions to find their influence. We will apply our algorithm to try to learn the whole model. Then use the resulting model to explain events through a Bayesian network or through the dynamics of the graph.

\subsubsection{Causal Bayesian Network learning}

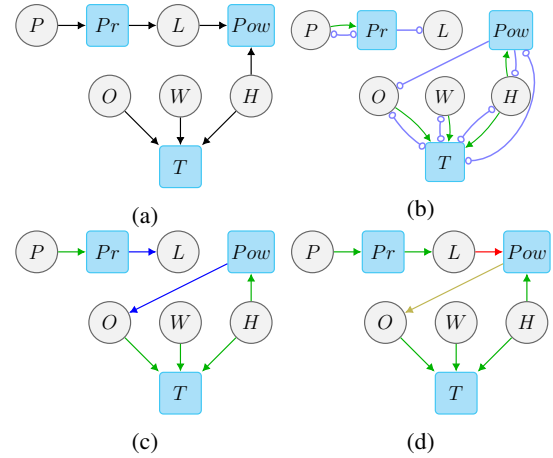
\begin{figure}[ht]
\centering
    \begin{subfigure}{0.4\linewidth}
    \centering  
    \resizebox{\linewidth}{!}{
    \begin{tikzpicture}[
    normal/.style={circle, draw=black!60, fill=black!5, very thick, minimum size=15mm},
    ND-node/.style={rectangle, rounded corners, draw=cyan!60, fill=cyan!30, very thick, minimum size=15mm},
    arrow/.style={-{Latex[width=3mm]}, very thick},
    ND-arrow/.style={-{o-onew, arrowhead=3mm}}
    ]
    %Nodes
    \node[normal,text centered]       (person)     at (0,0)        {\huge $P$};
    \node[ND-node,text centered, right=of person]       (presence)      {\huge $Pr$};
    \node[normal,text centered, right=of presence]      (light)  {\huge $L$};
    \node[ND-node,text centered, right=of light]       (power)    {\huge $Pow$};
    \node[normal,text centered, below=of power]       (heater)       {\huge $H$};
    \node[normal,text centered, left=of heater]       (window)       {\huge $W$};
    \node[normal,text centered, left=of window]       (outdoor)       {\huge $O$};
    \node[ND-node,text centered, below= of window]  (thermometer)   {\huge $T$};

    %Lines
    \draw[arrow] (person) -- (presence);
    \draw[arrow] (presence) -- (light);
    \draw[arrow] (light) -- (power);
    \draw[arrow] (heater) -- (power);
    \draw[arrow] (outdoor) -- (thermometer);
    \draw[arrow] (heater) -- (thermometer);
    \draw[arrow] (window) -- (thermometer);

    \end{tikzpicture}
    }
    \caption{}
    \label{fig:gt}
    \end{subfigure}
    \begin{subfigure}{0.4\linewidth}
    \centering  
    \resizebox{\linewidth}{!}{
    \begin{tikzpicture}[
            normal/.style={circle, draw=black!60, fill=black!5, very thick, minimum size=15mm},
            ND-node/.style={rectangle, rounded corners, draw=cyan!60, fill=cyan!30, very thick, minimum size=15mm},
            arrow/.style={-{Latex[width=3mm]}, very thick},
            ND-arrow/.style={-{Circle[open,width=3mm]},ultra thick, blue!50},
            double-ND-arrow/.style={{Circle[open,width=3mm]}-{Circle[open,width=3mm]},ultra thick, blue!50},
            missed/.style={-{Latex[width=3mm]}, very thick, red},
            correct/.style={-{Latex[width=3mm]}, very thick, black!30!green},
            flagged/.style={-{Latex[width=3mm]}, very thick, blue},
            added/.style={-{Latex[width=3mm]}, very thick, black!30!yellow},
        ]
        %Nodes
        \node[normal,text centered]       (person)     at (0,0)        {\huge $P$};
        \node[ND-node,text centered, right=of person]       (presence)      {\huge $Pr$};
        \node[normal,text centered, right=of presence]      (light)  {\huge $L$};
        \node[ND-node,text centered, right=of light]       (power)    {\huge $Pow$};
        \node[normal,text centered, below=of power]       (heater)       {\huge $H$};
        \node[normal,text centered, left=of heater]       (window)       {\huge $W$};
        \node[normal,text centered, left=of window]       (outdoor)       {\huge $O$};
        \node[ND-node,text centered, below= of window]  (thermometer)   {\huge $T$};

        %correct
        \draw[correct] (person) to[bend left=10] (presence);
        \draw[ND-arrow] (presence) -- (light);
        
        \draw[correct] (heater) to[bend left=10] (power);
        \draw[correct] (outdoor) to[bend left=10] (thermometer);
        \draw[correct] (heater) to[bend left=10] (thermometer);
        \draw[correct] (window) to[bend left=10] (thermometer);
        
        %added
        \draw[ND-arrow] (power) -- (outdoor);
        \draw[double-ND-arrow] (presence) to[bend left=10] (person);
        \draw[double-ND-arrow] (thermometer) to[bend left=10] (heater);
        \draw[double-ND-arrow] (thermometer) to[bend left=10] (window);
        \draw[double-ND-arrow] (thermometer) to[bend left=10] (outdoor);
        \draw[double-ND-arrow] (power) to[bend left=60] (thermometer);
        \draw[ND-arrow] (power) to[bend left=10] (heater);
    \end{tikzpicture}
    }
    \caption{}
    \label{fig:raw}
    \end{subfigure}
    \begin{subfigure}{0.4\linewidth}
        \centering
        \resizebox{\linewidth}{!}{
        \begin{tikzpicture}[
            normal/.style={circle, draw=black!60, fill=black!5, very thick, minimum size=15mm},
            ND-node/.style={rectangle, rounded corners, draw=cyan!60, fill=cyan!30, very thick, minimum size=15mm},
            arrow/.style={-{Latex[width=3mm]}, very thick},
            ND-arrow/.style={-{Circle[open,width=3mm]},ultra thick, blue},
            missed/.style={-{Latex[width=3mm]}, very thick, red},
            correct/.style={-{Latex[width=3mm]}, very thick, black!30!green},
            flagged/.style={-{Latex[width=3mm]}, very thick, blue},
            added/.style={-{Latex[width=3mm]}, very thick, black!30!yellow},
            ]
            
            % Nodes
            \node[normal,text centered]       (person)     at (0,0)        {\huge $P$};
            \node[ND-node,text centered, right=of person]       (presence)      {\huge $Pr$};
            \node[normal,text centered, right=of presence]      (light)  {\huge $L$};
            \node[ND-node,text centered, right=of light]       (power)    {\huge $Pow$};
            \node[normal,text centered, below=of power]       (heater)       {\huge $H$};
            \node[normal,text centered, left=of heater]       (window)       {\huge $W$};
            \node[normal,text centered, left=of window]       (outdoor)       {\huge $O$};
            \node[ND-node,text centered, below= of window]  (thermometer)   {\huge $T$};

            %correct
            \draw[correct] (person) -- (presence);
            \draw[flagged] (presence) -- (light);
            
            \draw[correct] (heater) -- (power);
            \draw[correct] (outdoor) -- (thermometer);
            \draw[correct] (heater) -- (thermometer);
            \draw[correct] (window) -- (thermometer);
            
            %added
            \draw[flagged] (power) -- (outdoor);
            
        \end{tikzpicture}
        }
        \caption{}
        \label{fig:clean}
    \end{subfigure}
    \begin{subfigure}{0.4\linewidth}
        \centering
        \resizebox{\linewidth}{!}{
        \begin{tikzpicture}[
            normal/.style={circle, draw=black!60, fill=black!5, very thick, minimum size=15mm},
            ND-node/.style={rectangle, rounded corners, draw=cyan!60, fill=cyan!30, very thick, minimum size=15mm},
            arrow/.style={-{Latex[width=3mm]}, very thick},
            ND-arrow/.style={-{Circle[open,width=3mm]},ultra thick, blue},
            missed/.style={-{Latex[width=3mm]}, very thick, red},
            correct/.style={-{Latex[width=3mm]}, very thick, black!30!green},
            flagged/.style={-{Latex[width=3mm]}, very thick, blue},
            added/.style={-{Latex[width=3mm]}, very thick, black!30!yellow},
            ]
            %Nodes
            \node[normal,text centered]       (person)     at (0,0)        {\huge $P$};
            \node[ND-node,text centered, right=of person]       (presence)      {\huge $Pr$};
            \node[normal,text centered, right=of presence]      (light)  {\huge $L$};
            \node[ND-node,text centered, right=of light]       (power)    {\huge $Pow$};
            \node[normal,text centered, below=of power]       (heater)       {\huge $H$};
            \node[normal,text centered, left=of heater]       (window)       {\huge $W$};
            \node[normal,text centered, left=of window]       (outdoor)       {\huge $O$};
            \node[ND-node,text centered, below= of window]  (thermometer)   {\huge $T$};

            %correct
            \draw[correct] (person) -- (presence);
            \draw[correct] (presence) -- (light);
            \draw[missed] (light) -- (power);
            \draw[correct] (heater) -- (power);
            \draw[correct] (outdoor) -- (thermometer);
            \draw[correct] (heater) -- (thermometer);
            \draw[correct] (window) -- (thermometer);
            
            %added
            \draw[added] (power) -- (outdoor);
            
        \end{tikzpicture}
        }
        \caption{}
        \label{fig:comp}
    \end{subfigure}
    \caption{Results of our approach applied to the smart home simulation. Fig \ref{fig:gt} shows Groundtruth causal model for the living room. Non-doable variables are shown in blue. The raw output of conditional testings, shown in (\ref{fig:raw}), is then processed to remove less significant arrows to obtain a DAG (\ref{fig:clean}). (\ref{fig:comp}): comparison between this output and the ground truth diagram from fig. \ref{fig:gt}: the red arrow is a missed relation while the yellow one is a connection wrongly added to the model.}
    \label{fig:results}
\end{figure}

The construction of the causal graph is shown in fig. \ref{fig:results}. First, observations of correlated variables and results of interventions yields a ``raw'' output depicted in fig \ref{fig:raw}. Note how the presence of ND-nodes introduces ND-arrows emerging from them. The next step of our algorithm processes this raw output to remove the least significant arrows to make it a DAG that is compatible with the observations. The output of this step, shown in fig. \ref{fig:clean}, contains two arrows flagged as ND-arrows. When comparing this final output to ground truth, in fig. \ref{fig:comp}, we notice that one of these ND-arrows was erroneous, displaying a performance limit in the case of ND-nodes. On the other hand, one causal relation, between light status and power consumption, was missed by our approach. This mistake can be explained in this situation, by the relatively small impact of light, in comparison to the heater, on power consumption.

%%%%%%% FIGURE a -> b
% Dans notre cas ici on n'a pas eu de cycle. En gros le test d'indépendance conditionnelle n'a pas supprimer la connection entre heater et le thermomètre il ne sont pas indépendant (cas 3). entre light et presence on a le cas 4. Enfin entre power <- heater -> thermomètre on a le cas 6

% The first step of our algorithm will return what we saw in figure \ref{fig:raw}. It shows the raw result after the successive tests to eliminate the connections. Then we will do the postprocessing which will apply the rules that we present in the previous section and return the final result (figure \ref{fig:results}). figure \ref{fig:comp} compare the result with the real causal graph, there we saw that we have found all the connections except the one between light and power. This is explained by the fact that the light has a very small influence on the power consumed. Finally, we have one additional connection between the power and the outdoor. This relationship is of type ND so it had to be removed by the conditional independence test and this one found that the two variables were correlated which avoided the removal of the connection.

%%%%  comp %%%%

Building on the structure of the causal graph presented in fig. \ref{fig:results}, we complete the learning process by using maximum likelihood estimates to finally provide a Causal Bayesian Network. 

\subsubsection{Impact of ND-nodes}

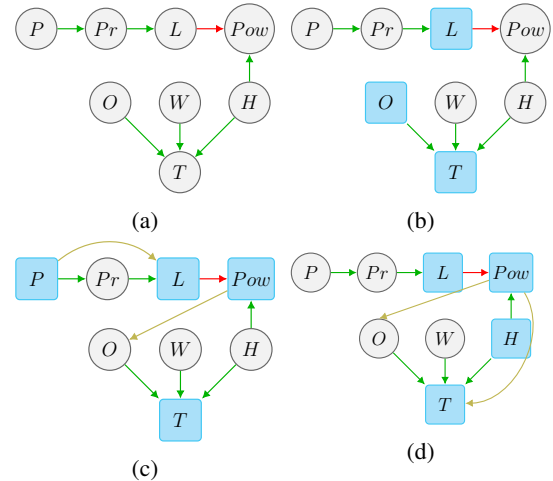
\begin{figure}
    \centering
    \begin{subfigure}{0.4\linewidth}
        \centering
        \resizebox{\linewidth}{!}{
        \begin{tikzpicture}[
            normal/.style={circle, draw=black!60, fill=black!5, very thick, minimum size=15mm},
            ND-node/.style={rectangle, rounded corners, draw=cyan!60, fill=cyan!30, very thick, minimum size=15mm},
            arrow/.style={-{Latex[width=3mm]}, very thick},
            ND-arrow/.style={-{Circle[open,width=3mm]},ultra thick, blue},
            double-ND-arrow/.style={{Circle[open,width=3mm]}-{Circle[open,width=3mm]},ultra thick, blue!50},
            missed/.style={-{Latex[width=3mm]}, very thick, red},
            correct/.style={-{Latex[width=3mm]}, very thick, black!30!green},
            flagged/.style={-{Latex[width=3mm]}, very thick, blue},
            added/.style={-{Latex[width=3mm]}, very thick, black!30!yellow},
            ]
            %Nodes
            \node[normal,text centered]       (person)     at (0,0)        {\huge $P$};
            \node[normal,text centered, right=of person]       (presence)      {\huge $Pr$};
            \node[normal,text centered, right=of presence]      (light)  {\huge $L$};
            \node[normal,text centered, right=of light]       (power)    {\huge $Pow$};
            \node[normal,text centered, below=of power]       (heater)       {\huge $H$};
            \node[normal,text centered, left=of heater]       (window)       {\huge $W$};
            \node[normal,text centered, left=of window]       (outdoor)       {\huge $O$};
            \node[normal,text centered, below= of window]  (thermometer)   {\huge $T$};

            %correct
            \draw[correct] (person) -- (presence);
            \draw[correct] (presence) -- (light);
            \draw[missed] (light) -- (power);
            \draw[correct] (heater) -- (power);
            \draw[correct] (outdoor) -- (thermometer);
            \draw[correct] (heater) -- (thermometer);
            \draw[correct] (window) -- (thermometer);

        \end{tikzpicture}
        }
        \caption{}
        \label{fig:}
    \end{subfigure}
    \begin{subfigure}{0.4\linewidth}
        \centering
        \resizebox{\linewidth}{!}{
        \begin{tikzpicture}[
            normal/.style={circle, draw=black!60, fill=black!5, very thick, minimum size=15mm},
            ND-node/.style={rectangle, rounded corners, draw=cyan!60, fill=cyan!30, very thick, minimum size=15mm},
            arrow/.style={-{Latex[width=3mm]}, very thick},
            ND-arrow/.style={-{Circle[open,width=3mm]},ultra thick, blue},
            double-ND-arrow/.style={{Circle[open,width=3mm]}-{Circle[open,width=3mm]},ultra thick, blue!50},
            missed/.style={-{Latex[width=3mm]}, very thick, red},
            correct/.style={-{Latex[width=3mm]}, very thick, black!30!green},
            flagged/.style={-{Latex[width=3mm]}, very thick, blue},
            added/.style={-{Latex[width=3mm]}, very thick, black!30!yellow},
            ]
            %Nodes
            \node[normal,text centered]       (person)     at (0,0)        {\huge $P$};
            \node[normal,text centered, right=of person]       (presence)      {\huge $Pr$};
            \node[ND-node,text centered, right=of presence]      (light)  {\huge $L$};
            \node[normal,text centered, right=of light]       (power)    {\huge $Pow$};
            \node[normal,text centered, below=of power]       (heater)       {\huge $H$};
            \node[normal,text centered, left=of heater]       (window)       {\huge $W$};
            \node[ND-node,text centered, left=of window]       (outdoor)       {\huge $O$};
            \node[ND-node,text centered, below= of window]  (thermometer)   {\huge $T$};

            %correct
            \draw[correct] (person) -- (presence);
            \draw[correct] (presence) -- (light);
            \draw[missed] (light) -- (power);
            \draw[correct] (heater) -- (power);
            \draw[correct] (outdoor) -- (thermometer);
            \draw[correct] (heater) -- (thermometer);
            \draw[correct] (window) -- (thermometer);

        \end{tikzpicture}
        }
        \caption{}
        \label{fig:}
    \end{subfigure}
    \begin{subfigure}{0.4\linewidth}
        \centering
        \resizebox{\linewidth}{!}{
        \begin{tikzpicture}[
            normal/.style={circle, draw=black!60, fill=black!5, very thick, minimum size=15mm},
            ND-node/.style={rectangle, rounded corners, draw=cyan!60, fill=cyan!30, very thick, minimum size=15mm},
            arrow/.style={-{Latex[width=3mm]}, very thick},
            ND-arrow/.style={-{Circle[open,width=3mm]},ultra thick, blue},
            missed/.style={-{Latex[width=3mm]}, very thick, red},
            correct/.style={-{Latex[width=3mm]}, very thick, black!30!green},
            flagged/.style={-{Latex[width=3mm]}, very thick, blue},
            added/.style={-{Latex[width=3mm]}, very thick, black!30!yellow},
            ]
            %Nodes
            \node[ND-node,text centered]       (person)     at (0,0)        {\huge $P$};
            \node[normal,text centered, right=of person]       (presence)      {\huge $Pr$};
            \node[ND-node,text centered, right=of presence]      (light)  {\huge $L$};
            \node[ND-node,text centered, right=of light]       (power)    {\huge $Pow$};
            \node[normal,text centered, below=of power]       (heater)       {\huge $H$};
            \node[normal,text centered, left=of heater]       (window)       {\huge $W$};
            \node[normal,text centered, left=of window]       (outdoor)       {\huge $O$};
            \node[ND-node,text centered, below= of window]  (thermometer)   {\huge $T$};

            %correct
            \draw[correct] (person) -- (presence);
            \draw[correct] (presence) -- (light);
            \draw[missed] (light) -- (power);
            \draw[correct] (heater) -- (power);
            \draw[correct] (outdoor) -- (thermometer);
            \draw[correct] (heater) -- (thermometer);
            \draw[correct] (window) -- (thermometer);
            
            %added
            \draw[added] (person) to[bend left=40] (light);
            \draw[added] (power) -- (outdoor);
        \end{tikzpicture}
        }
        \caption{}
        \label{fig:}
    \end{subfigure}
    \begin{subfigure}{0.4\linewidth}
        \centering
        \resizebox{\linewidth}{!}{
        \begin{tikzpicture}[
            normal/.style={circle, draw=black!60, fill=black!5, very thick, minimum size=15mm},
            ND-node/.style={rectangle, rounded corners, draw=cyan!60, fill=cyan!30, very thick, minimum size=15mm},
            arrow/.style={-{Latex[width=3mm]}, very thick},
            ND-arrow/.style={-{Circle[open,width=3mm]},ultra thick, blue},
            double-ND-arrow/.style={{Circle[open,width=3mm]}-{Circle[open,width=3mm]},ultra thick, blue!50},
            missed/.style={-{Latex[width=3mm]}, very thick, red},
            correct/.style={-{Latex[width=3mm]}, very thick, black!30!green},
            flagged/.style={-{Latex[width=3mm]}, very thick, blue},
            added/.style={-{Latex[width=3mm]}, very thick, black!30!yellow},
            ]
            %Nodes
            \node[normal,text centered]       (person)     at (0,0)        {\huge $P$};
            \node[normal,text centered, right=of person]       (presence)      {\huge $Pr$};
            \node[ND-node,text centered, right=of presence]      (light)  {\huge $L$};
            \node[ND-node,text centered, right=of light]       (power)    {\huge $Pow$};
            \node[ND-node,text centered, below=of power]       (heater)       {\huge $H$};
            \node[normal,text centered, left=of heater]       (window)       {\huge $W$};
            \node[normal,text centered, left=of window]       (outdoor)       {\huge $O$};
            \node[ND-node,text centered, below= of window]  (thermometer)   {\huge $T$};

            %correct
            \draw[correct] (person) -- (presence);
            \draw[correct] (presence) -- (light);
            \draw[missed] (light) -- (power);
            \draw[correct] (heater) -- (power);
            \draw[correct] (outdoor) -- (thermometer);
            \draw[correct] (heater) -- (thermometer);
            \draw[correct] (window) -- (thermometer);

            %added
            \draw[added] (power) -- (outdoor.north);
            \draw[added] (power) to[bend left=60] (thermometer);
        \end{tikzpicture}
        }
        \caption{}
        \label{fig:}
    \end{subfigure}
    
    \caption{Causal graphs obtained with different configuration of non-doable nodes (in blue). Comparison with the ground-truth relation (\ref{fig:gt}: green arrows are correct, while red, yellow and blue respectively denote missed, wrongly added or bidirectional connections.}
    \label{fig:newsetup}
\end{figure}

To analyze the impact of ND-nodes on the learning process, we have explored different arrangements of ND-nodes, by allowing or not variable modifications in the simulation. Figure \ref{fig:newsetup} shows the resulting graphs our algorithm learns, depending on the do-ability of nodes. Overall, relations between non-doable nodes are affected, while the algorithm stays robust concerning the other relations in the graph. 

\subsubsection{Unusual Causal Relations}

A motivation for learning CBN with minimal prior knowledge was the ability to adapt to unusual situations, detect them and use knowledge to provide explanation to the user. Such a situation may occur in our experimental setup: the thermometer has been implemented to be sensitive to the heat produced by nearby devices (notably the light or the heater). In our setup, four rooms are simulated with the same devices, however the precise location of devices within each room is random. This leads to situations where, in one of the rooms, the light is sufficiently close to the light so that a new causal relation $(light \rightarrow thermometer)$ appears in the causal model of the room. Faced with this event, our algorithm was able to see the new connection in the corresponding room.

\begin{figure}
    \centering
        \resizebox{0.7\linewidth}{!}{
        \begin{tikzpicture}[
            normal/.style={circle, draw=black!60, fill=black!5, very thick, minimum size=15mm},
            ND-node/.style={rectangle, rounded corners, draw=cyan!60, fill=cyan!30, very thick, minimum size=15mm},
            arrow/.style={-{Latex[width=3mm]}, very thick},
            ND-arrow/.style={-{Circle[open,width=3mm]},ultra thick, blue},
            missed/.style={-{Latex[width=3mm]}, very thick, red},
            correct/.style={-{Latex[width=3mm]}, very thick, black!30!green},
            flagged/.style={-{Latex[width=3mm]}, very thick, blue},
            added/.style={-{Latex[width=3mm]}, very thick, black!30!yellow},
            ]
            %Nodes
            \node[normal,text centered]       (light1)   at (-6,0) {\huge $L_1$};
            \node[ND-node,text centered]  (thermometer1)  at (-2,0) {\huge $T_1$};
            
            \node[normal,text centered]       (light2)   at (2,0) {\huge $L_2$};
            \node[ND-node,text centered]  (thermometer2)  at (6,0) {\huge $T_2$};
            
            \node[inner sep=0pt] (russell) at (0,5)
                {\includegraphics[width=1.5\linewidth]{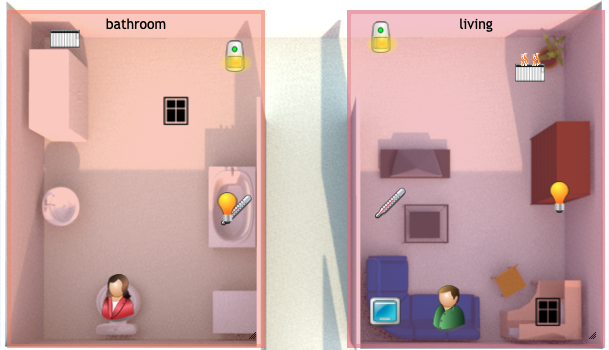}};
            %correct
            \draw[correct] (light1) -- (thermometer1);
            \draw[very thick, red] (-2,4.2) circle (1) ;
            
        \end{tikzpicture}
        }
    
    \caption{ In the bathroom, the light is close enough to the thermometer to influence temperature measures. This relation is detected by our algorithm (below).}
    \label{fig:lighteffect}
\end{figure}
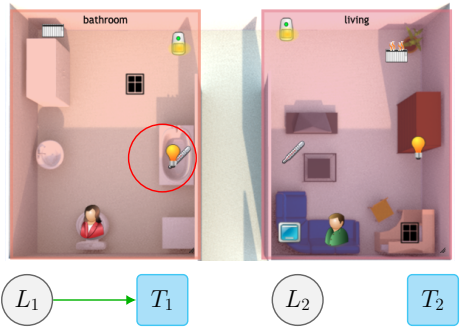

In this particular case, the diagnostic inference offered by our approach allows finding the cause of a particular behavior of the system (figure \ref{fig:lighteffect}): \emph{``The thermometer reports a hot temperature in the bathroom while the heater is off''}. In this case, diagnostic inference on the Bayesian network would initiate $T=1$ and $H=0$, and would infer the state of the lamp $P(L=1\mid (T,H)=(1,0))=0.61$.

A final line of analysis would be to look at the evolution of the causal graph over time. Its dynamics could allow us to explain abnormal phenomena that will occur after it has changed.

% We can obtain explanations about the phenomena in our environment by observing the evolution of our causal model. In our simulation, the thermometer is sensitive to the distance from objects that emit heat. So we will move the bulb closer to the thermometer and see what effect this will have on our causal model. As a result our algorithm was able to capture that and return a graph with an arrow going from light to temperature. 

% We could also compare the causal model of rooms between them to see if we have the same relationship between the variables from one to another. If this is not the case, we can analyze what differentiates them in order to draw conclusions. For the previous example we could not have this relationship between light and temperature in another room.  The difference between the two is the distance between the two objects. Thus, this offers us elements in our diagnosis to understand a potential malfunction related to the temperature measurements.

\section{Discussion and future work}
\label{sec:discussion}

While our experimental setup of the smart home was set to be close to a real-life use case, some limitations remain in our approach, some of which will be discussed here.

The first major assumption is that the system's causal model can be represented by an acyclic graph. This limitation is common in the literature on Causality Theory (\cite{peters_elements_2017, pearl_causality_2009, spirtes_causation_2000}. However, especially in the context of SAS, retro-actions may occur between devices and the variables they monitor: for instance, consider how a ``smart'' heater would turn on depending on the room's temperature. One potential workaround would be to take time into account, which removes any ambiguity regarding the direction of a causal relation. This can be done by learning a \emph{Temporal Bayesian Network} \cite{TBN}, which takes into account the time dependency.

However, adding time to the equation is no easy task: in sec. \ref{sec:experiments}, we argued that for our demonstration, we used a fixed time $\Delta_t$ after which we consider the consequences of interventions. This fixed value entails several issues: since different causal mechanisms can have different time characteristics, how can one know how long is long enough when waiting for the consequences of intervention? For future work, we may therefore consider implementing a multi-scale approach \cite{horstemeyer_multiscale_2009} to the environment model to cope with the possible different characteristic times.

% Commented this out as it doesn't seem that interesting
% Existing methods propose to estimate the time interval following interventions \cite{todo}: in the near future, we consider integrating a similar approach to the learning process of our Causal Bayesian Networks, as to reduce the number of parameters.

Furthermore, our approach requires a certain number of interventions on the system and has shown to perform better when only a limited number of variables are non-doable. These issues are minor in a simulated example but can be limiting when operating in a real-life environment. A workaround is to consider having access to a model on the environment, such as a ``digital twin''\cite{rosen_about_2015}, which our algorithm can use.

Having the causal diagram of a system, as opposed to a simple Bayesian Network, offers possibilities to be integrated into explanation frameworks. For instance, we may use Causal Bayesian Network in conjunction with the general Explanatory Engine proposed by \cite{houze_decentralized_2020}. This use case would benefit from both inference directions: diagnostic can be used as a powerful tool to propose hypotheses for abductive inference (i.e. finding the cause of an observed phenomenon), while predictive inference might be used to explore the potential consequences of a proposed solution, in the context of an explanation. The Causal Bayesian Network can also be converted into a Fuzzy Cognitive Map (FCM) which is used to present causal knowledge intuitively because of its simplicity. \cite{FCM} propose a method to generates an FCM from the bayesian network using Pearson’s correlation coefficient equation. However, here we can instead use the scores of the causality test instead of the correlation.

Future work may also focus on learning for large networks, for example, use multi-scale learning. We mean by that learning the model of rooms and then going to a larger scale to learn the interactions between the rooms.

\section{Conclusion}
\label{sec:conclusion}
We work towards the implementation in real-life SAS of the methods of Causality Theory. We have seen how intervention operations could be performed on a digital twin of an environment to train a causal diagram, which can later be used as a basis for our Causal Bayesian Network. This workflow has shown encouraging results in the example of smart home and, since it required no ad hoc knowledge about the particularities of the smart home context, can be generalized to other comparable setups.

Knowing the Causal Bayesian Graph offers advantages for applications such as explanations, given the more ``natural'' source of relations it entails, compared to a more traditional Bayesian Network. As such, we consider using this tool as a means to perform abductive and predictive inference in a broader explanatory framework for SAS.


%\section*{Acknowledgment}




%\bibliographystyle{plain}
%\bibliography{main.bib}

\printbibliography

\end{document}